\pdfoutput=1

\documentclass[11pt]{article}

\usepackage[final]{acl}

\usepackage{times}
\usepackage{latexsym}

\usepackage[T1]{fontenc}

\usepackage[utf8]{inputenc}

\usepackage{microtype}

\usepackage{inconsolata}

\usepackage{microtype}
\usepackage{hyperref}       
\usepackage{url}            
\usepackage{booktabs}       
\usepackage{amsfonts}       
\usepackage{nicefrac}       
\usepackage{microtype}      
\usepackage{amsmath}
\usepackage{amssymb}
\usepackage{mathtools}
\usepackage{amsthm}
\usepackage{graphicx}
\usepackage{booktabs} 
\usepackage{graphicx}
\usepackage{listings}
\usepackage{subcaption}
\usepackage{caption}
\usepackage{diagbox}
\usepackage{algorithm}
\usepackage{algpseudocode}
\usepackage{xcolor}         
\usepackage{hyperref}
\usepackage{tikz}
\usepackage{bbm}
\usepackage{pifont}
\usepackage{titletoc}
\usepackage{newfloat}

\DeclareFloatingEnvironment[
  fileext=lob,
  listname={List of Boxes},
  name=Prompt,
  placement=htb!,
]{PROMPT}



\title{Context-Aware Reasoning On Parametric Knowledge \\for Inferring Causal Variables} 

\author{%
  Ivaxi Sheth$^{1}$, Sahar Abdelnabi$^{2}$, Mario Fritz$^{1}$ \\ 
 $^{1} $CISPA Helmholtz Center for Information Security, $^{2}$ Microsoft \\
\texttt{\{ivaxi.sheth,fritz\}@cispa.de, saabdelnabi@microsoft.com}
}

\begin{document}
\maketitle
\begin{abstract}

Scientific discovery catalyzes human intellectual advances, driven by the cycle of hypothesis generation, experimental design, evaluation, and assumption refinement. Central to this process is causal inference, uncovering the mechanisms behind observed phenomena. While randomized experiments provide strong inferences, they are often infeasible due to ethical or practical constraints. However, observational studies are prone to confounding or mediating biases.
While crucial, identifying such backdoor paths is expensive and heavily depends on scientists' domain knowledge to generate hypotheses. 
We introduce a novel benchmark where the objective is to complete a partial causal graph. We design a benchmark with varying difficulty levels with over 4000 queries. We show the strong ability of LLMs to hypothesize the backdoor variables between a cause and its effect.  
Unlike simple knowledge memorization of fixed associations, our task requires the LLM to reason according to the context of the entire graph\footnote{Code available at \url{https://github.com/ivaxi0s/inferring-causal-variables} }.

\end{abstract}

\section{Introduction}

Scientific discovery has been key to humankind's advances. It is a dynamic process revolving around inquiry and refinement. Scientists adhere to a process that involves formulating a hypothesis and then collecting pertinent data~\cite{wang2023scientific}. They then draw inferences from these experiments, modify the hypothesis, formulate sub-questions, and repeat the process until the research question is answered~\cite{kiciman2023causal}. 

{Central to scientific discovery is formulating hypotheses and identifying relevant variables that drive the underlying causal mechanisms of observed phenomena~\cite{bunge2017causality}. Randomized controlled trials are the gold standard for establishing causal relationships, but they are often infeasible due to ethical, financial, or logistical constraints~\cite{nichol2010challenging}. In such cases, researchers rely on observational data, where a key challenge lies not only in analyzing relationships but in determining which variables should be observed and included in the analysis, particularly confounders or mediators that influence causal mechanisms underlying the outcomes~\cite{ananth2017confounding,gupta2021estimating}.}

\begin{figure}
    \vspace{-3mm}
    \centering
    \includegraphics[width=0.3 \textwidth]{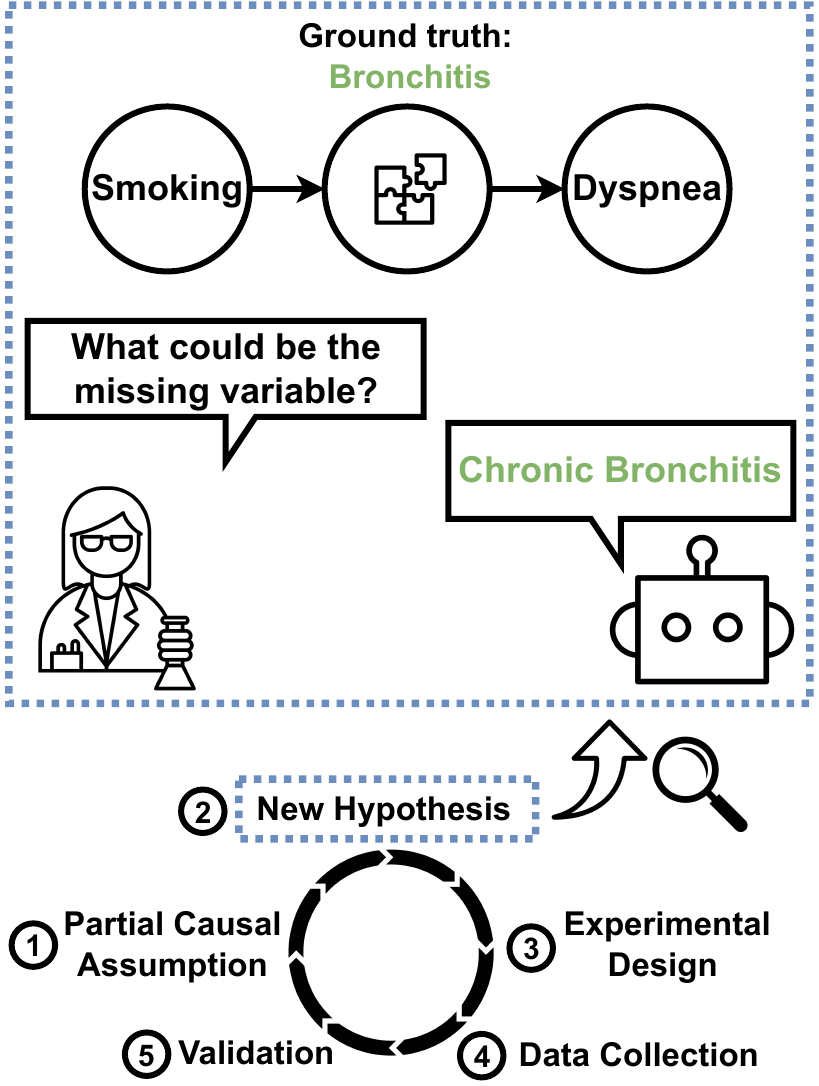}
    \caption{Scientific discovery iteratively generates hypotheses from assumptions using human expertise. We use LLMs as proxy experts to propose new hypotheses about missing variables in causal DAGs.}
    \label{fig:teaser}
    \vspace{-3mm}
\end{figure} 
With the recent advancement of Large Language Models (LLMs), there has been a growing interest in using them for scientific discovery~\cite{ai4science2023impact, lu2024ai, cory2024evolving}.
LLMs have demonstrated strong performance in internalizing knowledge~\cite{sun2024head,yukola} and reasoning-based tasks~\cite{valmeekam2023planbench, guo2025deepseek}, including causal discovery, where they infer pairwise causal relationships based on variable semantics~\cite{kiciman2023causal, long2023can, ban2023query, vashishtha2023causal, darvariu2024large, binkyte2025causality}.

Scientific reasoning is fundamentally context-driven; unlike simple factual retrieval, it requires adapting hypotheses based on new evidence and integrating knowledge across varying subpopulations. While recent work has explored the use of LLMs for causal discovery~\cite{kiciman2023causal, long2023can, darvariu2024large, ban2023query, vashishtha2023causal}, much of it assumes a fixed set of variables and focuses on identifying relationships among them. However, a critical and underexplored aspect is determining \emph{which} variables should be considered in the first place. This demands flexible, context-sensitive reasoning to identify missing causal factors.

In our paper, we use the term reasoning operationally to describe the model’s ability to generate hypotheses or identify variables that complete partial causal graphs. Our usage follows previous work by ~\citet{kiciman2023causal} in causal discovery and LLM research, where “reasoning” often refers to generating plausible hypotheses or prioritizing potential candidates given partial structural information, rather than strict deductive logic.

To address this gap, we propose a novel task: given a partial causal graph with missing variables, the LLM is prompted to hypothesize what those variables might be, using the structure and known nodes as context. By systematically omitting different variables, we generate diverse test cases to evaluate the robustness of model reasoning. We further decompose the benchmark into subtasks, starting from baseline variable identification to more realistic, open-ended settings where multiple unobserved mediators exist between known treatments and outcomes.

{Our task mirrors real-world scientific workflows, where identifying missing variables—especially confounders and mediators is essential for valid causal inference. This typically demands costly, interdisciplinary effort. LLMs, trained on diverse knowledge sources, offer a scalable alternative. For example, in a stroke drug study, an LLM might suggest socioeconomic status as an unmeasured confounder. While recent works advocate using LLMs as co-pilots for causal tasks~\cite{petersen2024artificial, alaa2024large}, systematic evaluations are lacking. Our benchmark addresses this gap by assessing LLMs' ability to infer missing causal variables across domains.}


Our main \textbf{contributions} are: 1) We propose and formalize the novel task of LLM-assisted causal variable inference. 2) We propose a benchmark for inferring missing variables across diverse domains of causal graphs. 3) We design experimental tasks with different difficulty levels and knowledge assumptions, such as open-world and closed-world settings, the number of missing variables, etc. 4) Our benchmark allows for both grounded evaluations and a reproducible framework to benchmark LLMs' capabilities in hypothesis generation.

\vspace{-1mm}
\section{Related Work}
\paragraph{LLMs and Causality.}
Our work builds on the foundational framework of causality by~\citet{pearl2009causality}. Prior studies have explored extracting causal relationships from text~\cite{girju2002text, hassanzadeh2020causal, tan2023unicausal, dhawan2024end} and using LLMs for causal reasoning~\cite{kiciman2023causal}, including commonsense~\cite{frohberg2021crass, singh2021com2sense} and temporal causality~\cite{zhang2020reasoning, zhang2022rock}. Recent efforts prompt LLMs with variable names to discover causal structures~\cite{kiciman2023causal,long2023can,darvariu2024large,ban2023query,vashishtha2023causal}. Others integrate LLMs with deep structural causal models~\cite{anonymous2024causal, yu2019dag}, or focus on graph formatting~\cite{sheth2024causalgraph2llm}, query design~\cite{jiralerspong2024efficient}, and causal inference~\cite{jin2023can}.
In contrast to prior work, we use LLMs to infer missing variables before data collection and evaluation, leveraging their pre-trained knowledge for this novel hypothesizing task. 
\paragraph{LLMs and Hypothesis Generation.}
Existing work tested hypothesis generation with LLMs in reasoning tasks or free-form scientific hypotheses from background knowledge provided in the context~\cite{gendron2023large,qi2023large,xu2023large, xu2023llms, qiu2023phenomenal, lu2024ai}. In contrast, we consider the structured task of causal hypothesis generation, where the ground-truth variables are known and can be used for evaluation. 
\paragraph{Context-aware reasoning} has been explored through prompt engineering~\cite{97994e4526ef7eeea59190aa466fbab05fad9187,zhou2022least, ranaldi2023hans}, premise ordering manipulation~\cite{chen2024premise}, diagnostic analyses~\cite{prabhakar2024deciphering}, and compositional reasoning evaluations~\cite{press2022measuring,saparov2024testing}. Unlike premise-based or linguistic evaluations, our setup requires reasoning over causal graph topology, using contextual cues by varying assumptions.

\section{Preliminaries: Causal Graph}
\label{sec:variabledesc}
A causal relationship can be modeled via a Directed Acyclic Graph (DAG). A causal DAG represents relationships between a set of $N$ variables defined by \(\mathbf{{V}} = \{ v_1,...,v_N \}\). The variables are encoded in a graph $\mathcal{G} = (\mathbf{{V}}, \mathbf{{E}})$ where $\mathbf{\textrm{E}}$ is a set of directed edges between the nodes $\in \mathbf{\textrm{V}}$ such that no cycle is formed. Mathematically, it can be expressed as:
\vspace{-1mm}
 \[\mathcal{G} = (\mathbf{{V}}, \mathbf{{E}}),\]
 \vspace{-7mm}
  \[\mathbf{{E}} = \{e_{i,j} \mid v_i, v_j \in \mathbf{{V}} , i \neq j \enspace \text{and} \: v_i \rightarrow v_j \}\]
Each edge $e_{i,j}$ denotes causal relationship and the influence from $v_i$ to $v_j$, \(v_i \xrightarrow{e_{i,j}} v_j\). 

We define $\mathbf{d}(v)$ as the degree of a node $v$, representing the total number of edges connected to $v$. 
$\mathbf{d_{\text{in}}}(v)$ is the in-degree, representing the number of incoming edges to $v$. 
$\mathbf{d_{\text{out}}}(v)$ is the out-degree, representing the number of outgoing edges from $v$.

\noindent\textbf{Source} has no incoming edges; \(d_{\text{in}}(v) = 0\). \\
\textbf{Sink} has no outgoing edges. Sinks are \(d_{\text{out}}(v) = 0\). \\
\textbf{Treatment} is characterized by nodes that are being intervened upon. \\
\textbf{Outcome} is characterized by nodes that are observed for interventions from the treatments. \\
\textbf{Mediator} has both incoming and outgoing edges (\(d_{\text{in}}(v) > 0\) and \(d_{\text{out}}(v) > 0\)) as intermediaries in the pathways between treatment and outcome. \\
\textbf{Confounder} influences both treatment and outcome, exhibiting edges directed towards the treatment and outcome nodes (\(d_{\text{out}}(v) \geq 2 ) \). Hence $v$ is a confounder if it is a parent of both \(v_i\) and \(v_j\).\\
\textbf{Collider} has two edges meeting, and \(d_{\text{in}}(v) > 1\). I.e., $v$ is a collider if it is a child of both \(v_i\) and \(v_j\). 
\section{Inferring Causal Variables}

Motivated by the challenge of discovering variables that block backdoor paths to ensure unbiased causal inference~\cite{glymour2019review}, in this work, we leverage language models to infer missing variables in a causal DAG. We assume that a part of the graph is already known, and the aim is to find additional variables that can be incorporated into the existing DAG to enhance the underlying causal mechanism. 

Formally, we assume a partially known causal DAG, $\mathcal{G}^* = ({\textrm{V}^*}, \mathbf{\textrm{E}})$, where ${\textrm{V}}^* \subseteq {\mathbf{V}}$. The objective is to identify the set of missing variables ${\textrm{V}^*} = {\mathbf{V}} \setminus {\textrm{V}}_{\text{missing}}$ thereby expanding $\mathcal{G}^*$ to $\mathcal{G}$. This implies that all causal relationships (edges) among variables in $V^*$ are known and correctly represented in $\mathcal{G}^*$; i.e., $\textrm{E}$ is fully specified. Here, ``missing'' variables are not latent or hidden by measurement error but known unknowns within the causal graph reflective of the LLM's perspective.

To systematically assess LLMs' ability to infer missing causal variables, we construct a multi-stage benchmark with increasing levels of complexity.  We begin with a controlled setting, where the model is provided with a partial causal DAG and a set of multiple-choice options to identify missing variables. 
Then, the task becomes open-ended, where LLMs hypothesize missing variables, simulating an open-world paradigm. Additionally, as the task escalates, we introduce more complexity by omitting additional nodes, challenging the model to hypothesize multiple missing variables. 

We evaluate the reasoning capability of LLMs through prompting. We represent the graph $\mathcal{G}^*$ using a prompt template $P_{\text{LLM}}(\cdot)$ which enables LLMs to parse causal relationships in the DAG.

\subsection{Task 1: Out-of-Context Identification}

\label{subsec:4.1}
\textit{Motivation.} To assess whether LLMs can infer missing variables in causal graphs, we begin with a controlled multiple-choice setting that serves as a baseline. This task isolates the core challenge: identifying a single missing variable from a causal DAG.
By restricting the search space to a fixed set of options, including the correct variable and out-of-context distractors, we evaluate whether the model can distinguish the variable that meaningfully completes the causal structure. 

The partial DAG $\mathcal{G}^*$ is created by removing one variable, denoted as $v_x$, from the original DAG $\mathcal{G}$. 
The role of the LLM is to select a variable from the multiple choices, $\text{MCQ}_{v_x}$, that can be used to complete the graph.
\begin{figure*}
   \begin{subfigure}[t]{0.45\textwidth}
         \centering
         \includegraphics[width=0.8\textwidth]{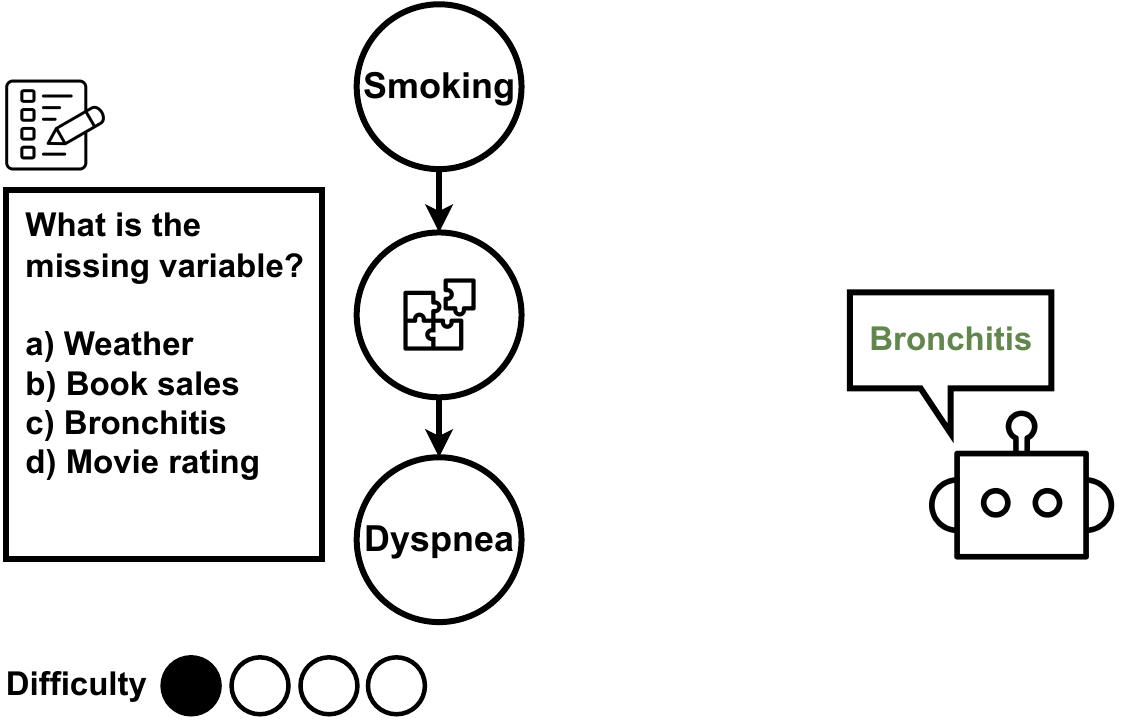}
         \caption{Task 1}
         \label{fig:diff1}
     \end{subfigure}
   \begin{subfigure}[t]{0.45\textwidth}
         \centering
         \includegraphics[width=0.8\textwidth]{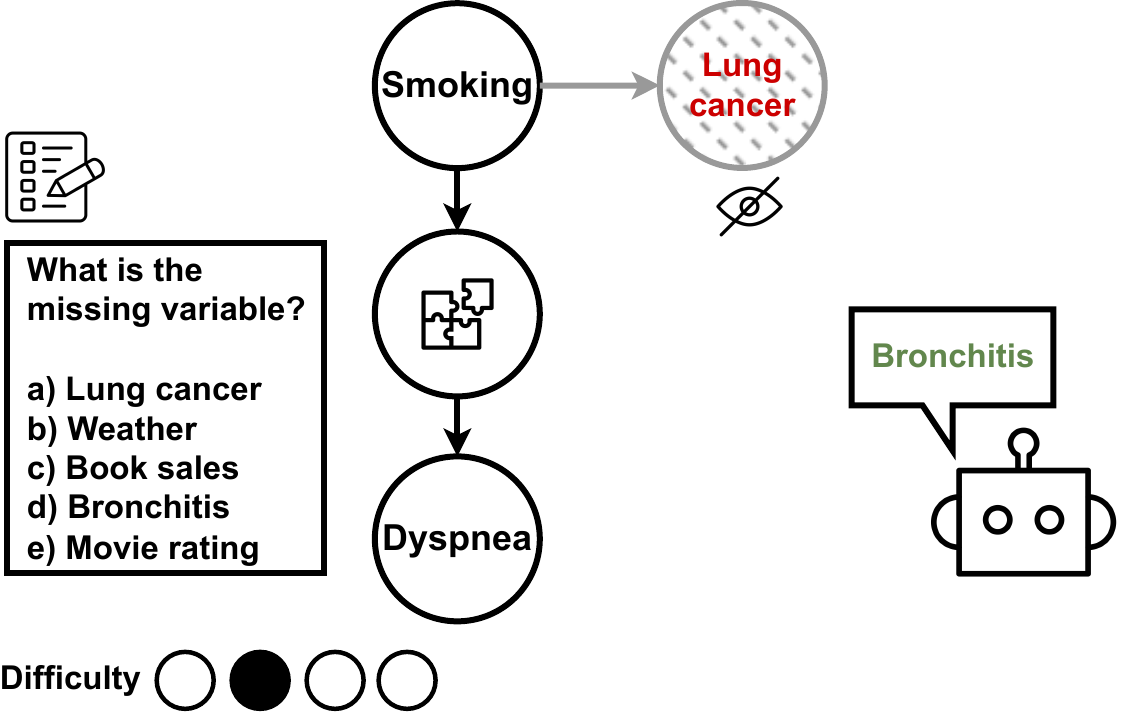}
         \caption{Task 2}
         \label{fig:diff2}
     \end{subfigure}
     \caption{Leveraging LLM to identify the missing variable for a causal DAG in the presence of out-of-context distractors (a), an in-context distractor along with out-of-context distractors (b).}
\vspace{-1mm}
\end{figure*}  
The out-of-context distractors are unrelated to the causal domain of the given DAG, chosen to minimize any contextual overlap with the true missing variable. Let $v_x^*$ represent the variable selected by the LLM to complete $\mathcal{G}^*$.
 \[v_x^* = P_{\text{LLM}}(\mathcal{G}^*, \text{MCQ}_{v_x}) \: \: \: \forall  v_x \in \textbf{{V}}\]
 
\subsection{Task 2: In-Context Identification}
\textit{Motivation.}
In real-world domains like healthcare and finance, missing or unobserved variables often challenge causal inference~\cite{hughes2019accounting,tian2012testable}. This task simulates such ambiguity by requiring LLMs to identify a relevant missing variable when presented with multiple plausible options, going beyond the baseline.

Here, instead of removing one node from the ground truth DAG $\mathcal{G}$, two nodes, $v_{x_1}$ and $v_{x_2}$, are now removed to create the partial graph, $\mathcal{G}^{*}$.
\[\mathcal{G}^{*} = \mathcal{G} \setminus \{v_{x_1}, v_{x_2}\} \quad \text{for} \quad v_{x_1}, v_{x_2} \in \mathbf{V}\]
The MCQA paradigm provides multiple choices, including the missing variables $v_{x_1}$ and $v_{x_2}$. The task for the LLM here is to select the correct variable $v_{x_1}$ only, given an in-context choice $v_{x_2}$ and out-of-context choices. The in-context variables are plausible within the same causal graph, allowing the LLM to use DAG-defined context inference to distinguish the relevant from the irrelevant options.
We ensure $v_{x_1}$ and $v_{x_2}$ are not directly connected i.e., neither is a parent of the other. 
\begin{align*}
    v_{x_1}^{*} &= P_{\text{LLM}}(\mathcal{G^*}, \text{MCQ}_{v_{x_1}, v_{x_2}})  \:\:\:
    \forall \: v_{x_1}, v_{x_2} \in \mathbf{V}  \:   \:  
\end{align*}
\vspace{-7mm}
\begin{align*}
    \text{and} \: v_{x_1} \not\rightarrow v_{x_2}, \:\: v_{x_2} \not\rightarrow v_{x_1} 
\end{align*}
\subsection{Task 3: Hypothesizing in Open World}
\textit{Motivation.}
Previous tasks constrained the model to select from predefined options. However, real-world reasoning rarely offers such scaffolding. This task increases complexity by removing the multiple-choice format entirely.

Given a partial DAG $\mathcal{G}^*$, formed by removing a node $v_x$, the model must generate potential missing variables without any provided candidates (see~\autoref{fig:diff3}). The output is a ranked list of hypotheses $\{v_{x,1}^{*},..., v_{x,k}^{*}\}$ for $k$ suggestions, simulating open-ended discovery.
\vspace{-1mm}
\[ \{v_{x,1}^{*}, v_{x,2}^{*}, ..., v_{x,k}^{*}\} =P_{\text{LLM}}(\mathcal{G}^{*}) \: \forall \: v_{x} \in \mathbf{V}\]

\subsection{Task 4: Iteratively Hypothesizing in Open World}
\begin{figure*}
   \begin{subfigure}[t]{0.45\textwidth}
         \centering
         \includegraphics[width=0.8\textwidth]{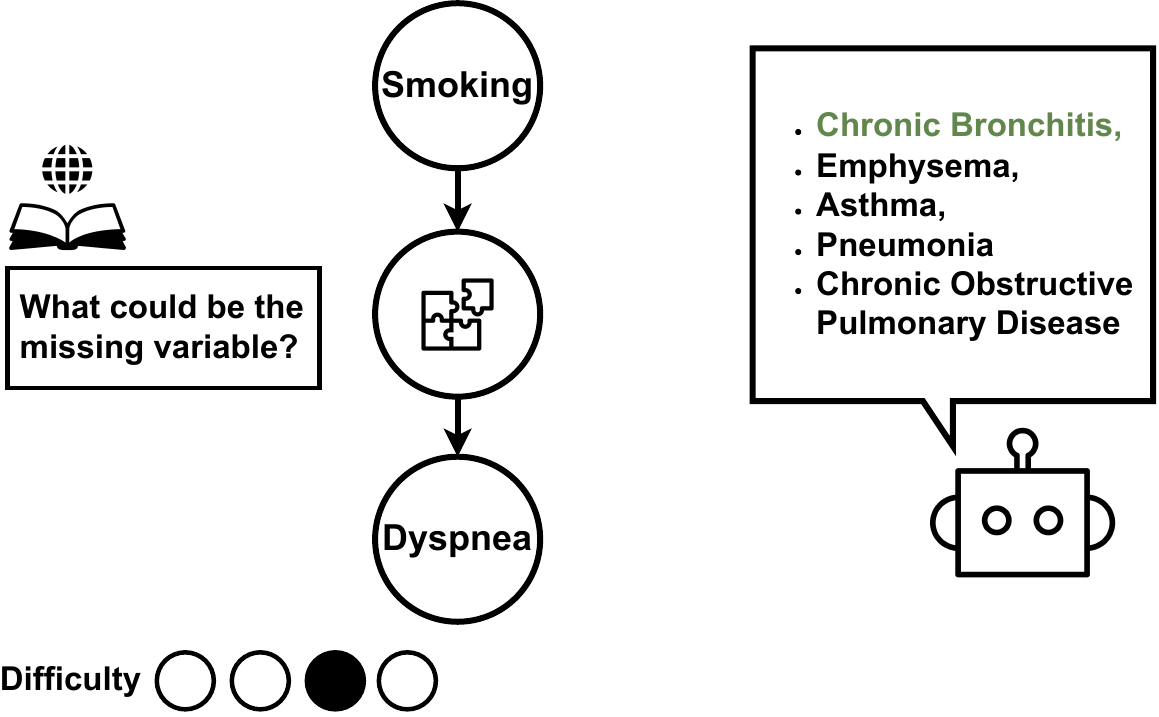}
         \caption{Task 3}
         \label{fig:diff3}
     \end{subfigure}
   \begin{subfigure}[t]{0.45\textwidth}
         \centering
         \includegraphics[width=0.8\textwidth]{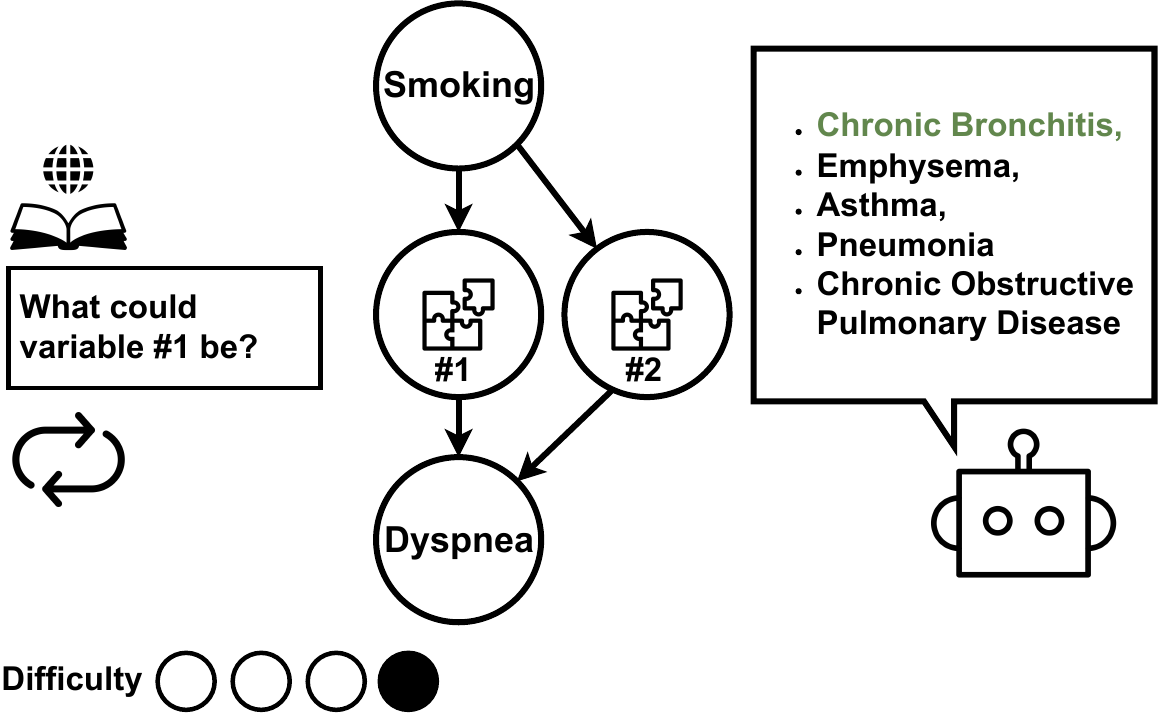}
         \caption{Task 4}
         \label{fig:diff4}
     \end{subfigure}
     \caption{Leveraging LLM to hypothesize the missing variable in a causal DAG in an open-world setting for one variable (a), in an iterative fashion for multiple missing mediators (b). }
\vspace{-2mm}
\end{figure*}  
\textit{Motivation.} Building on the open-world setting, we further increase task difficulty by removing multiple nodes from the causal graph. The goal is no longer to recover a single missing variable but to iteratively hypothesize a set of mediators that link a treatment to an outcome.

Given a partial DAG $\mathcal{G}^* = \mathcal{G} \setminus {v_{x_1}, \ldots, v_{x_M}}$, the task (illustrated in~\autoref{fig:diff4}) involves generating a sequence of missing mediators $M = {v_{m_1}, v_{m_2}, ..., v_{m_H}}$ that plausibly connect a treatment variable $v_t$ to an outcome variable $v_y$.

At each iteration $i$, , the LLM is prompted with the current partial graph and returns a hypothesis for the next mediator. This process continues until all of the mediators are inferred.
\[   v_{m_i}^* = P_{\text{LLM}}(\mathcal{G}^* \cup \{v_{m_1}^*, ..., v_{m_{i-1}}^*\}),
\]
for $i = 1, ..., H$. The sequence of mediators $M = \{v_{m_1}, v_{m_2}, ..., v_{m_H}\}$ is chosen at random.

To assess how mediator order affects performance, we draw on mediation analysis concepts~\cite{pearl2014interpretation}, specifically the Natural Direct Effect (NDE)—the treatment’s effect not mediated by a variable—and the Natural Indirect Effect (NIE)—the portion mediated by it (see Appendix~\ref{app:nde}). We propose the Mediation Influence Score (MIS) to quantify each mediator's impact between a treatment and outcome. Defined as the ratio of NIE to NDE, MIS is a scale-free, positive measure of a mediator’s relative contribution:
\begin{align*}
    \text{MIS} \: (v_{m_i}) = \left| \frac{\text{NIE}(v_{m_i)}}{\text{NDE}(v_{m_i})}\right| \quad \text{for} \quad i = 1, ..., H.
\end{align*}
This metric quantifies the relative importance of the indirect effect (through the mediator) compared to the direct impact. 
Mediators are then ranked and prioritized based on their MIS scores, with higher scores indicating a stronger mediation effect.

\section{Evaluation and Results}


\textbf{Graphs.}
We evaluate a variety of causal graphs spanning diverse domains. We use the semi-synthetic DAGs from BNLearn repository - Cancer~\cite{korb2010bayesian}, Survey~\cite{scutari2021bayesian}, Asia~\cite{lauritzen1988local}, Child~\cite{spiegelhalter1992learning}, Insurance~\cite{binder1997adaptive}, and Alarm~\cite{beinlich1989alarm}. We also evaluate our approach on a realistic Alzheimer's Disease graph~\cite{anonymous2024causal}, developed by five domain experts and Law~\cite{vanderweele2011causal}. See Appendix~\ref{app:dataset} for further details. 

\begin{table}[htb!]
\footnotesize
    \centering
    \begin{tabular}{l|cc|p{3.8cm}}
    \toprule
       \textbf{Graph} & \textbf{$\mathbf{\textrm{V}}$} & \textbf{$\mathbf{\textrm{E}}$}  & \textbf{Description} \\ \midrule
        Cancer & $5$ & $4$ & Factors around lung cancer \\
        Survey & $6$ & $6$ & Factors for choosing transportation\\ 
        Asia & $8$ & $8$ & Factors affecting dysponea\\
        Law & $8$ & $20$ & factors around legal system \\
        Alzheimer & $9$ & $16$ & Factors around Alzheimer's Disease \\
        Child & $20$ & $25$ & Lung related illness for a child\\
        Insurance & $27$ & $52$ & Factors affecting car accident insurance \\
        Alarm & $37$ & $46$ & Patient monitoring system\\
    \midrule
    \end{tabular}
    \caption{Datasets used in the benchmark.}
    \label{tab:data_stats}
\end{table}

\noindent\textbf{Models.} We evaluate our setups across different open-source and closed models. The models we use are GPT-4o~\cite{hurst2024gpt}, GPT-4~\cite{openai2023gpt4}, LLama3-chat-8b~\cite{touvron2023Llama}, Mistral-7B-Instruct-v0.2~\cite{jiang2023mistral}, Mixtral-7B-Instruct-v0.1~\cite{jiang2024mixtral}, Zephyr-7b-Beta~\cite{tunstall2023zephyr} and Neural-chat-7b-v3-1~\cite{neuralchat}. 

\noindent \textbf{Prompt.} We used the textual prompting strategy from ~\citet{sheth2024causalgraph2llm} after performing experiments on some of the proposed encoding methods (see Appendix~\ref{app:encoding}).
Implementation details are in Appendix~\ref{app:implementation} and prompts in Appendix~\ref{app:prompt}. Our code will be available after anonymity period.

\subsection{Task 1} 
\noindent\textit{Setup.} The input to the LLM consists of a partial DAG $\mathcal{G}^*$, and multiple choices including the correct missing variable $v_x$ and several out-of-context distractors. This task includes 120 queries. We define accuracy to assess the LLM's $v_x$ prediction.
\vspace{-2mm}
 \[\text{Accuracy} = \frac{1}{N} \sum_{i=1}^{N} \mathbbm{1} (v_{x}^{*} = v_{x}^{i})\]
\vspace{-2mm}
\begin{figure*}
   \begin{subfigure}[t]{0.48\textwidth}
         \centering
         \includegraphics[width=0.8\textwidth]{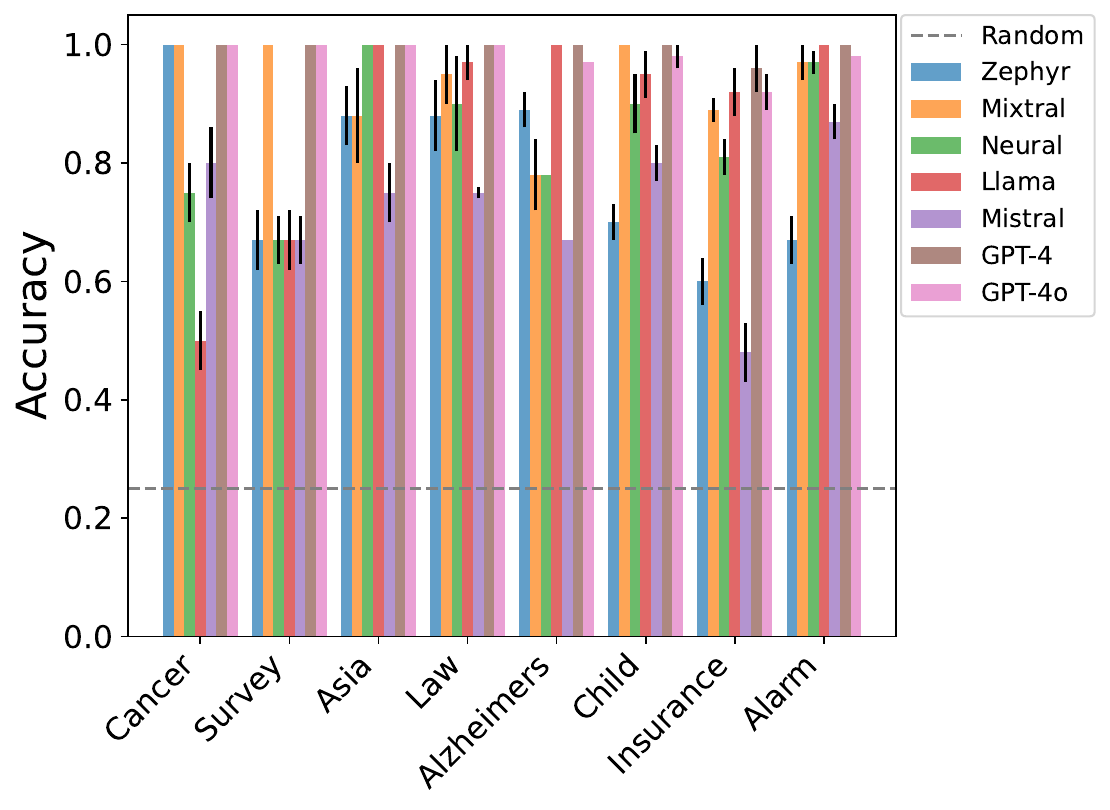}
         \vspace{-2mm}
         \caption{Task 1 Result.}
         \label{fig:mcq}
     \end{subfigure}
   \begin{subfigure}[t]{0.48\textwidth}
         \centering
         \includegraphics[width=0.8\textwidth]{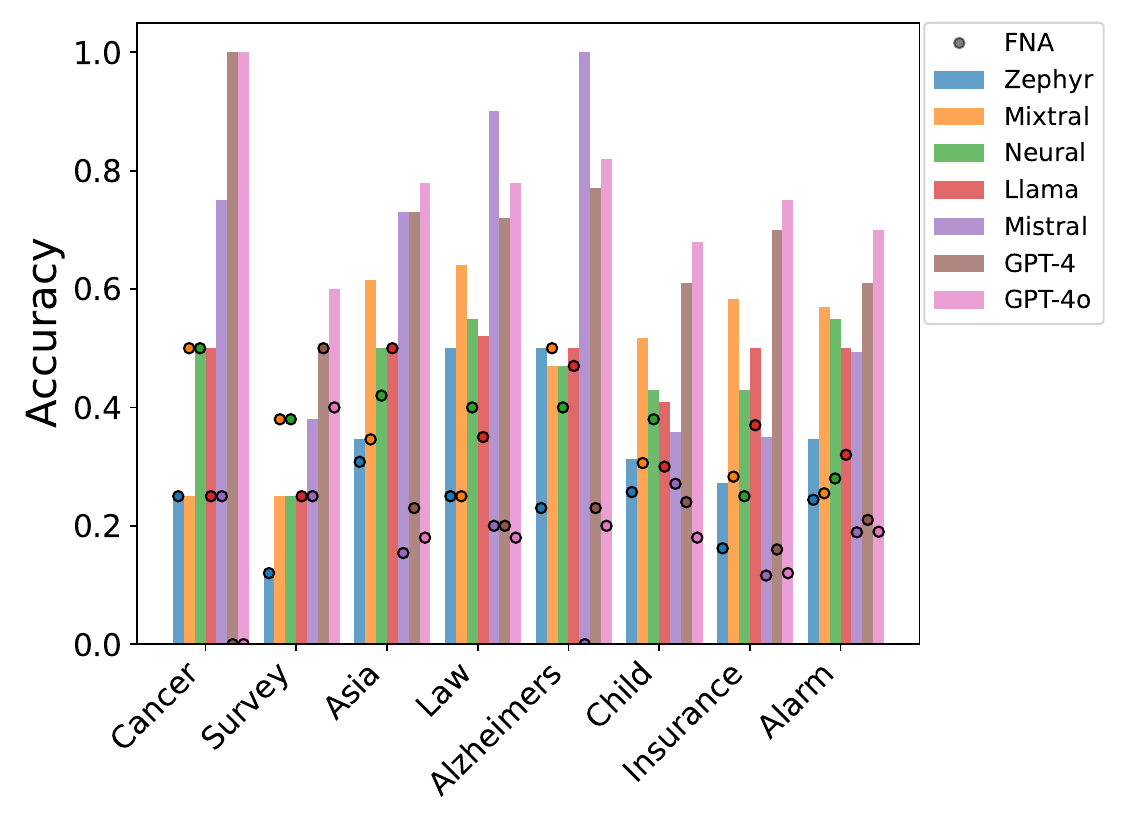}
         \vspace{-2mm}
         \caption{Task 2 Result.}
         \label{fig:2nodes}
     \end{subfigure}
    \caption{Accuracy of LLMs in identifying the missing causal variable from multiple choices with out-of-context distractors (a), and from both out-of-context and in-context distractors (b).} 
\vspace{-1mm}
\end{figure*}  

\noindent\textbf{Results.}
In~\autoref{fig:mcq}, we report the accuracy of different LLMs in identifying the missing variable. 
GPT-4, followed closely by Mixtral and GPT-4o, consistently performs well, achieving perfect accuracy on most of the graphs. Other models, including Mistral-7b, Llama-7b, Neural, and Zephyr-7b, have varying degrees of success. Insurance remains the most challenging graph, potentially due to the high number of edges present in the DAG. All models significantly outperform the random baseline. However, we conjecture that the high performance could be partially attributed to the simplicity of the task. The models might be using the context of the graph domain to exclude unrelated distractors rather than engaging in deeper causal reasoning among multiple plausible choices. To investigate this, we introduce an in-domain choice among the multiple choices in the next experiment.

\subsection{Task 2}
\textit{Setup.} This is a more challenging task where the partial graph has two missing nodes. In addition to out-of-context distractors and the ground-truth variable, $v_{x_1}$, the multiple-choice set includes the second missing variable $v_{x_2}$ as an in-context distractor. This setup tests the model’s ability to reason over indirect causal relations contextually to identify the correct variable. This task results in over 3800 queries. To evaluate performance, we use two metrics: Accuracy and False Node Accuracy (FNA). FNA captures how often the model incorrectly selects the in-context distractor instead:
\vspace{-2mm}
\[\text{FNA} \downarrow = \frac{1}{N} \sum_{i=1}^{N} \mathbbm{1}(v_{x_1}^{*} = v_{x_2})
\]
\vspace{-2mm}

\begin{table*}[tb!]
\small
\setlength{\tabcolsep}{1.5pt}
    \centering
    \begin{tabular}{l|cc|cc|cc|cc|cc|cc|cc|cc|cc}
    \toprule
        & \multicolumn{2}{c|}{{Cancer}} & \multicolumn{2}{c|}{{Survey}} & \multicolumn{2}{c|}{{Asia}} & \multicolumn{2}{c|}{{Law}}& \multicolumn{2}{c|}{{Alzheimers}} & \multicolumn{2}{c|}{{Child}} & \multicolumn{2}{c|}{{Insurance}} & \multicolumn{2}{c|}{{Alarm}}    & \multicolumn{2}{c}{{Avg}} \\  \cmidrule{2-19}
      & \tiny{Sim} & \tiny{LLM-J} & \tiny{Sim} & \tiny{LLM-J} & \tiny{Sim} & \tiny{LLM-J} & \tiny{Sim} & \tiny{LLM-J} & \tiny{Sim} & \tiny{LLM-J} & \tiny{Sim} & \tiny{LLM-J} & \tiny{Sim} & \tiny{LLM-J} & \tiny{Sim} & \tiny{LLM-J} & \tiny{Sim} & \tiny{LLM-J}\\\midrule

        Zephyr & ${0.36}$ & ${0.61}$ & ${0.34}$ & ${0.60}$ & ${0.45}$ & ${0.66}$ & $0.41$ & ${0.70}$ & ${0.35}$ & ${0.75}$ & ${0.51}$ & ${0.70}$ & ${0.45}$ & ${0.44}$ & ${0.46}$ & ${0.69}$  & ${0.42}$ & ${0.63}$\\
        
        Mixtral & $0.41$ & $0.66$ & $0.39$ & $0.66$ & $\mathbf{0.66}$ & $0.75$ & $0.38$ & $0.69$ & $0.31$ & $0.77$ & $\mathbf{0.53}$ & $\mathbf{0.77}$ & $0.46$ & ${0.56}$ & $\mathbf{0.50}$ & $0.72$ & $0.46$  & $0.70$ \\
        
        Neural & $0.38$ & $0.77$ & $0.43$ & $0.55$ & $0.53$ & $0.55$ & $0.47$ & $0.72$ & $0.44$ & $0.71$ & $0.48$ & $0.70$ & $0.47$ & $0.43$ & $0.47$ & $0.67$  & $0.45$  & $0.63$  \\
        
        Llama & $0.40$ & $0.48$ & $0.40$ & $0.54$ & $0.53$ & $0.58$ & $\mathbf{0.67}$ & $0.65$ & $0.45$ & $0.61$ & $0.48$ & $0.63$ & $0.42$ & $0.34$ & $0.46$ & $0.65$ & $0.45$  & $0.55$  \\
        
        Mistral & $0.33$ & $0.67$ & $0.44$ & $0.65$ & $0.60$ & $0.73$ & $0.49$ & $0.67$ & $0.34$ & $0.76$ & $0.48$ & $0.68$ & $0.46$ & $0.47$ & $0.47$ & $0.71$ & $0.44$  & $0.67$  \\
        
        
        GPT-4 & $\mathbf{0.49}$ & $\mathbf{0.90}$ & $\mathbf{0.51}$ & $0.67$ & $\mathbf{0.66}$ & ${0.76}$ & $0.55$ & ${0.78}$ & $0.47$ & $\mathbf{0.98}$ & $0.36$ & $0.53$ & $0.52$ & $0.56$ & $0.49$ & $0.75$ & $0.50$  & $0.73$  \\

        GPT-4o & $0.52$ & $0.89$ & $0.50$ & $\mathbf{0.71}$ & $\mathbf{0.66}$ & $\mathbf{0.78}$ & $0.58$ & $\mathbf{0.80}$ & $\mathbf{0.50}$ & $0.91$ & $0.40$ & $0.60$ & $\mathbf{0.54}$ & $\mathbf{0.58}$ & $0.44$ & $\mathbf{0.76}$ & $\mathbf{0.54}$ & $\mathbf{0.76}$ \\

    \bottomrule
    \end{tabular}
    \caption{Task 3 Results. Average semantic similarity and LLM-as-Judge metrics to evaluate LLMs in hypothesizing the missing variable in a causal DAG.}
    \label{tab:open_world}
    \vspace{-2mm}
\end{table*}
\noindent\textbf{Results.} In~\autoref{fig:2nodes}, we report Accuracy and False Node Accuracy (FNA) across graphs. Accuracy reflects how often the correct missing variable is chosen, while FNA measures how often the model incorrectly selects the in-context distractor—another missing variable included to test deeper causal reasoning. Since there are 5 options, random accuracy is $0.2$, and FNA under random guessing would be around $0.2$ as well. GPT-4 and GPT-4o achieve high accuracy and low FNA, showing that they reliably distinguish the true missing node from both distractors and the in-context variable. GPT-4o slightly outperforms GPT-4 on several graphs. Open models like Mistral, Zephyr, and Mixtral show more variability, performing well on simpler graphs like Cancer but struggling on complex ones like Alarm. While most models exceed random chance, higher FNA in some cases highlights a tendency to confuse plausible but incorrect variables, emphasizing the difficulty of reasoning over multiple missing nodes.

\subsection{Task 3}
\textit{Setup.}
In real-world settings, partial causal graphs provided by domain experts often lack ground truth and multiple choices. Hypotheses may vary depending on context, data, or domain knowledge. To simulate this, we prompt the LLM to generate. The LLM generates \(k=5\) suggestions for the missing node \(v_{x}\). This task has 120 queries. We compare suggestions to the ground truth, recognizing that real-world cases often lack a single correct answer. Since traditional metrics may miss contextual nuances, we use two evaluations: semantic similarity and LLM-as-Judge (see Appendix~\ref{app:llmjudgevs}). 
\begin{enumerate}
    \item \textbf{Semantic Similarity.} We compute the cosine similarity between the embeddings of the predictions, \(v_{x_{1:5}}^*\), and the ground truth \(v_{x}\), averaging the highest similarity scores across all nodes \(v_x \in \textbf{V}\) (see Appendix~\ref{app:semantic} for details). 

    \item \textbf{LLM-Judge.} Inspired by~\citet{zheng2023judging}, this two-step metric assesses contextual semantic similarity beyond exact matches. First, LLM ranks suggestions \(v_{x_{1:5}}^*\) based on how well they fit the partial graph. Second, it rates the best match on a 1–10 scale. Scores are averaged across nodes for an overall measure (see Appendix~\ref{app:sec:llmj}).  
\end{enumerate}

\begin{figure}[htb!]
\centering
     \begin{subfigure}[t]{0.35\textwidth} 
         \centering
         \includegraphics[width=\textwidth]{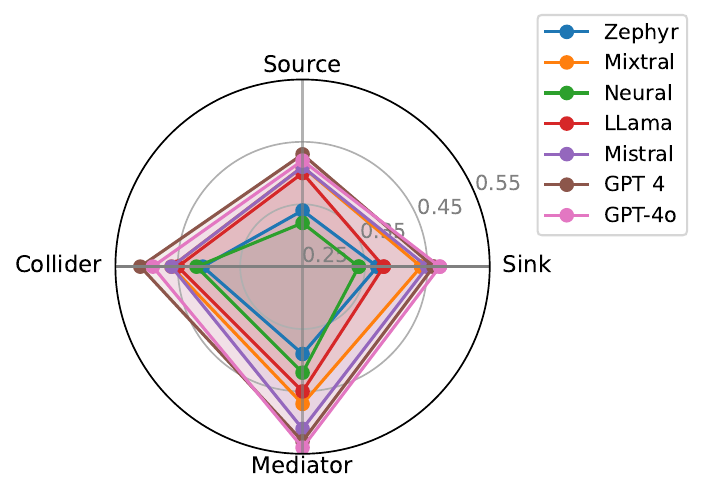} 
         \caption{Semantic similarity.}
     \end{subfigure}
     \begin{subfigure}[t]{0.35\textwidth}
         \centering
         \includegraphics[width=\textwidth]{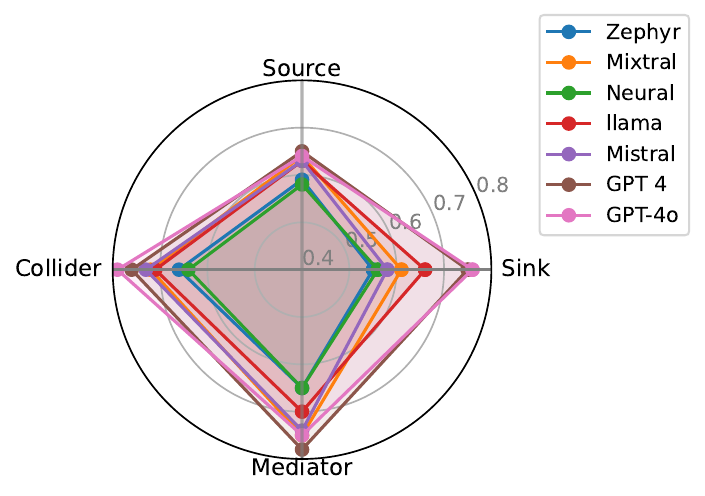}
         \caption{LLM-as-Judge.}
     \end{subfigure}
    \caption{Task 3 Results. Visualizing each model's performance, averaged across the different graphs, for Sink, Source, Mediator, and Collider nodes.}
    \label{fig:avg_spider}
    \vspace{-4mm}
\end{figure}

 \noindent\textbf{Results.}
We report models' performances using both semantic similarity and LLM-Judge metrics in ~\autoref{tab:open_world}. For brevity, we provided the variances in Appendix~\ref{app:tab:open_world}. 
We provide a detailed analysis of each metric across different types of node variables (defined in Section~\ref{sec:variabledesc}). We evaluate sources, sinks, colliders, and mediators for each of the partial causal graphs. The results, fine-grained by node type, are given in Figure~\ref{fig:avg_spider}, which shows each model's average performance across graphs with a detailed performance per graph in~\autoref{fig:spider_detail}.
GPT-4, GPT-4o and Mixtral generally achieve higher semantic similarity and LLM-as-Judge scores across most graphs (\autoref{fig:spider_detail}). We observe that semantic similarity is a stricter metric than LLM-as-judge since it cannot encode contextual information about the causal DAG (see example in~\autoref{tab:mainssvsllm}). Despite different scales, both metrics seem to be fairly correlated. 
\autoref{fig:avg_spider}, 
shows that models display stronger performance for colliders and mediators on average. This suggests that these models are better at reasoning about common causes and indirect causal relationships. 
Sinks are typically the nodes that represent the outcomes or effects of interventions (treatments) applied to other nodes. Source nodes represent the causes in a causal graph. Lower performance on these nodes indicates to reason about the potential causes and outcomes of the causal graphs is difficult. 

In~\autoref{fig:topk}, model performance improves with more suggestions ($k$). \autoref{fig:nodevsacc} shows that accuracy also correlates with node degree ($d_{in} + d_{out}$), indicating that more context aids prediction. Overall, LLMs perform well on many nodes, especially mediators and colliders, making them promising tools for real-world causal discovery where treatments and outcomes are known.

\subsubsection{Hypothesizing Confounder}
Backdoor paths are alternative causal pathways that confound the estimation of causal effects and introduce bias if not accounted for. Hence, hypothesizing and controlling for confounders is an important task in causal inference~\cite{pourhoseingholi2012control}. We extract confounder subgraphs from~\cite{sachs2005causal}, Alarm, and Insurance graphs. 
\begin{table}[htb!]
\vspace{-1mm}
\setlength{\tabcolsep}{10pt}
\small
    \centering
    
    \begin{tabular}{l|r|r|r}
    \toprule
       &{Sachs} & Alarm& Insurance \\ \midrule

Zephyr & $0.10$ & $0.45$ & $0.53$ \\ 
Mixtral & $\mathbf{0.95}$ & $\mathbf{0.85}$ & $0.63$ \\
Neural & $0.30$ & $0.45$ & $0.61$ \\
LLama & $0.20$ & $0.47$ & $0.63$ \\
Mistral & $0.20$ & $\mathbf{0.85}$ & $0.61$ \\
GPT-4 & $\mathbf{0.95}$ & $0.73$ & $\mathbf{0.78}$ \\

GPT-4o & $\mathbf{0.95}$	&$ 0.70$ & $0.73$ 
\\ \bottomrule

    \end{tabular}
    \caption{Hypothesizing Confounders in Task 3.}
    \label{tab:test_acc}
    
\end{table}
From~\autoref{tab:test_acc} and Appendix~\ref{app:confounders}, we find that while LLMs accurately hypothesize some confounders, models struggle with domain-specific graphs like SACHS. Larger models like GPT-4o don't necessarily always perform best, underscoring the need for diverse benchmarks. 

\subsection{Task 4}
\textit{Setup.} We adopt an iterative approach for hypothesizing mediators, allowing the model to refine predictions step-by-step—unlike global prediction, which yields lower performance (Appendix~\ref{app:iterativevsall}). This aligns with Chain-of-Thought~\cite{wei2022chain} reasoning and improves accuracy. There are more than 140 queries for this task, ranging from 1-10 missing mediators.
For \textbf{unordered evaluation}, mediators are given in random order and scored via average semantic similarity. For \textbf{ordered evaluation}, we rank mediators using the Mediation Influence Score (MIS) and compare model performance when prompted in ascending vs. descending MIS order. We define a metric, $\Delta$, to capture this difference.

\setlength{\tabcolsep}{3.4pt}
\begin{table}[htb!]
\tiny
    \centering
    \begin{tabular}{l|cc|cc|cc|cc}
    \toprule
       & \multicolumn{2}{c|}{Asia} & \multicolumn{2}{c|}{Child}  & \multicolumn{2}{c|}{Insurance} & \multicolumn{2}{c}{Alarm} \\ 
       & Sim & $\Delta$ & Sim & $\Delta$ & Sim & $\Delta$ & Sim & $\Delta$ \\ \midrule
        Zephyr & $0.61 $ & $-0.02 $ & $ \mathbf{0.54}$ & $\ \ \ \ 0.17$ & $ 0.47$ & $ \ \ \ \ 0.19$ & $0.51 $ & $ \ \ \ \ 0.20 $ \\
        Mixtral & $\mathbf{0.87} $ & $ \ \ \ \ 0.01 $ & $ 0.50$ &  $ \ \ \ \ 0.18$ & $ 0.48$ & $ \ \ \ \ 
 0.15 $ & $0.52 $ & $\ \ \ \ 0.13 $\\
        Neural & $ 0.65$ & $ \ \ \ \ 0.04$ & $ 0.48$ &  $ \ \ \ \ 0.21$ & $ 0.42$ & $ \ \ \ \ 0.16 $ & $ 0.46$ & $\ \ \ \ 0.12 $\\
        Llama & $0.80$ & $ \ \ \ \ 0.07$ & $ 0.49$ &  $-0.05$ & $ 0.44$ & $ \ \ \ \ 0.21$ & $ 0.51$  & $ \ \ \ \ 0.07$\\
        Mistral & $0.33$ & $ \ \ \ \ 0.02$ & $ 0.50$ & $ \ \ \ \ 0.12$ & $ 0.48$ & $ \ \ \ \ 0.13$ & $0.47 $ & $ \ \ \ \ 0.11$\\
        GPT-4 & $0.49$ & $ \ \ \ \ 0.04 $ & $ 0.39$ & $ \ \ \ \ 0.16$ & $ \mathbf{0.52}$ & $ \ \ \ \ 0.14$ & $ {0.60}$ & $-0.07$\\

        GPT-4o & $0.55$ & $ \ \ \ {0.00}$ & $0.48$ & $ \ \ \ {0.10}$ & $0.51$ & $ \ \ \ {0.08}$ & $\mathbf{0.62}$ & $ \ \ \ {0.01}$ \\

    \bottomrule
    \end{tabular}
    \caption{Task 4 Results. Accuracy of iterative mediator prediction when prompted in random order. $\Delta$ reflects the change in performance when mediators are ordered by their Mediation Influence Score (MIS). }
    \label{tab:iterM}
\end{table}

\noindent\textbf{Results.}
The results of this experiment are in~\autoref{tab:iterM}. Results with variances are provided in Appendix~\ref{app:tab:open_world}. In this highly complex environment with more than one node missing and with open-world search space, LLMs can still maintain their performance. 
Unlike the overall consistent performance of GPT-4 across all graphs, other models showed superior performance in Insurance and Alarm graphs only. As the complexity of the graph increases, we observe larger differences in hypothesizing the mediators according to the MIS order. Positive $\Delta$ values suggest that prompting the LLM based on the MIS metric leads to higher semantic similarity between the mediator hypotheses and the ground truth variables. In summary, we observe that LLMs can be effective in iteratively hypothesizing multiple mediators in a DAG, and if present, 
some domain knowledge about the significance of the mediator can boost the performance.

\subsection{Memorization}
{
A concern in evaluating pretrained LLMs on knowledge-intensive tasks is contamination i.e., memorization of evaluation data from training. This is especially relevant for public datasets like those in the BNLearn repository, which may have appeared in training corpora.}

{To assess this, we tested whether models could recall the number and names of variables from each of the eight datasets in our benchmark. This included well-known BNLearn graphs (e.g., Asia, Child, Insurance, Alarm) and less common ones (e.g., Law, Alzheimer's). We prompted each model to report node counts and variable names, including explicit references to BNLearn for relevant datasets, to detect signs of memorization. }

\begin{table}[htb]
\tiny
\centering
\begin{tabular}{lcccccccc}
\toprule
\textbf{Model} & \textbf{Cancer} & \textbf{Survey} & \textbf{Asia} & \textbf{Law} & \textbf{Alz} & \textbf{Child} & \textbf{Insurance} & \textbf{Alarm} \\
\midrule
Zephyr     & \ding{55} & \ding{55} & \ding{55}   & \ding{55} & \ding{55}   & \ding{55}   & \ding{55}   & \ding{55}    \\
Mixtral    & \ding{55} & \ding{55} & 0.71 & \ding{55} & \ding{55}   & \ding{55}   & \ding{55}   & 0.13 \\
Neural     & \ding{55} & \ding{55} & \ding{55}   & \ding{55} & \ding{55}   & \ding{55}   & \ding{55}   & \ding{55}    \\
LLama      & \ding{55} & \ding{55} & \checkmark   & \ding{55} & \ding{55}   & \ding{55}   & \ding{55}   & \ding{55}    \\
Mistral    & \checkmark & \ding{55} & \ding{55}   & \ding{55} & \ding{55}   & \ding{55}   & \ding{55}   & \ding{55}    \\
GPT-4      & \checkmark & \checkmark & \checkmark   & \ding{55} & 0.55 & \checkmark & \checkmark & \checkmark \\
GPT-4o    & \checkmark & \checkmark & \checkmark   & \ding{55} & 0.45 & \checkmark & \checkmark & \checkmark   \\

\bottomrule
\end{tabular}
\caption{Memorization analysis: Whether the model could correctly recall node information from the dataset (\checkmark), failed to recall (\ding{55}), or proportion of nodes recalled.}
\label{tab:contamination}
\vspace{-2mm}
\end{table}

{In~\autoref{tab:contamination}, except GPT family models, which exhibited partial recall for some widely known BNLearn datasets, we observe that full reconstruction of the graphs' details was rare. This recall was consistently absent for lesser-known datasets such as Law and Alzheimer's, which are less likely to have appeared during pretraining. While these findings cannot eliminate memorization with certainty, they suggest that it is not predominant for most models.}

{To further test GPT-4, we explicitly mentioned the graph provenance (e.g., ``This graph is from BNLearn'') during ``Task 3'', shown in~\autoref{tab:bnlearn_context_effect}. GPT-4’s performance improved across most graphs. This suggests that its initial responses were not purely reciting these graphs but potentially based on broader parametric knowledge.}



\subsection{Discussion}
The results show that LLMs effectively hypothesize missing variables, especially mediators, though performance varies with task complexity. Simple tasks, like identifying missing variables from controlled options, had high success rates. 

Performance differences across domains may stem from biases in LLM training data, affecting parametric memory. For instance, confounder hypothesis quality varied across graphs, with domain-specific gaps lowering accuracy, like in the Sachs graph (Appendix~\ref{app:confounders}).

We explored fine-tuning and few-shot prompting to enhance performance, but small DAG sizes limited the graph size, yielding mixed results (Appendix~\ref{app:ft}). While fine-tuning may help specialization, it can also reduce reliance on general parametric knowledge~\cite{yang-etal-2024-unveiling}. Future work could explore domain-specific fine-tuning.

Though model training data is undisclosed, we used a recently released graph~\cite{anonymous2024causal} that postdates cut-off dates (at the time of performing experiments). Our novel task and verbalization approach further reduce the risks of memorization. Table~\ref{tab:open_world} confirms LLMs generate novel hypotheses rather than retrieving memorized patterns, with no evidence of direct graph reconstruction. Our work relies on reasoning via parametric knowledge rather than explicit memorization. 

Our setup assumes known edges among missing variables for controlled evaluation, which future work can extend. We envision this as a human-LLM collaboration under expert supervision, as LLMs cannot self-assess plausibility or confidence~\cite{zhou2024relying}. Future work could also refine filtering mechanisms and improve performance on source and sink nodes.

\subsection{Human Evaluation}
To complement our automatic evaluation metrics, we conducted a small-scale human evaluation on three representative graphs (Cancer, Survey, and Asia). Two independent annotators (a CS PhD student and a CS PhD graduate) rated the quality of LLM-suggested variables. We then measured the agreement between human judgments, semantic similarity, and our LLM-judge using Spearman correlation.

\begin{table}[htb]
\small
\centering
\begin{tabular}{l|c|c}
\toprule
\textbf{} & {Correlation} & {p-value} \\
\midrule
Sim -- LLM-judge    & 0.430 & 0.2475 \\
Sim -- R1           & 0.781 & 0.0130 \\
Sim -- R2           & 0.623 & 0.0732 \\
LLM-judge -- R1     & 0.622 & 0.0738 \\
LLM-judge -- R2     & 0.831 & 0.0055 \\
R1 -- R2            & 0.823 & 0.0065 \\
\bottomrule
\end{tabular}
\caption{Spearman correlations between human annotators (R1, R2), semantic similarity (Sim), and LLM-judge scores on Cancer, Survey, and Asia graphs.}
\label{tab:human_eval}
\end{table}

\autoref{tab:human_eval} indicates strong correlations among human annotators and between human judgments and the automatic metrics. In particular, the LLM-judge shows high alignment with both annotators, suggesting that it serves as a reliable proxy for human evaluation. This supports the use of our automatic evaluation framework as a scalable approach for benchmarking causal reasoning tasks.

\section{Conclusion}

Most causality research focuses on identifying relationships from observed data, while hypothesizing which variables to observe remains largely reliant on expert knowledge. We propose using LLMs as proxies for this step and introduce a novel task: hypothesizing missing variables in causal graphs. We formalize this with a benchmark that spans varying levels of difficulty and ground-truth knowledge. 
Our results highlight LLMs' strengths in inferring backdoor paths, including colliders, confounders, and mediators, which often lead to biased causal inference when unaccounted for. Our work LLMs can serve as useful tools for early-stage hypothesis generation, supporting scientists in formulating plausible causal variables before data collection. By evaluating models across different graph completeness, open- and closed-world settings, we highlight their potential and limitations.

\section{Limitations}
While this work presents promising advancements in leveraging LLMs for hypothesizing missing variables in causal graphs, there are some limitations to consider. Our evaluation relies on established DAGs and comparisons with known ground truth, limiting assessment in scenarios without a defined baseline. Future work can include validation using human in loop evaluation. Future work can also integrate our work into the full causal discovery pipeline with statistical data.

\section{Ethics and Risk}
Our work leverages LLMs for hypothesis generation in causal discovery but comes with ethical risks. Biases from training data may lead to skewed hypotheses, and over-reliance on AI without expert validation could result in misleading conclusions. While we design our task to minimize memorization, risks of data leakage remain. Additionally, LLM performance varies across domains, making errors in high-stakes fields like healthcare particularly concerning. To mitigate these risks, we emphasize human-AI collaboration, transparency in model limitations, and improved evaluation frameworks for reliability.

\section{Acknowledgements}
This work was partially funded by ELSA – European Lighthouse on Secure and Safe AI funded by the European Union under grant agreement No. 101070617 and the German Federal Ministry of Education and Research (BMBF) under the grant AIgenCY (16KIS2012). Views and opinions expressed are however those of the authors only and do not necessarily reflect those of the European Union or European Commission. Neither the European Union nor the European Commission can be held responsible for them. 

\bibliography{main}
\appendix

\newpage
\appendix
\onecolumn

\section{Implementation}
\label{app:implementation}
\subsection{Datasets}
\label{app:dataset}
We use 7 real-world based graphs. These graphs span different domain knowledge topics. These graphs have ground truth graphs along with their observational data. The simplest graph used is the cancer graph with 4 edges and 5 node variables. In addition to the semi-synthetic graphs from the BNLearn library, we also evaluate our approach on a realistic Alzheimer's Disease graph~\cite{anonymous2024causal}, which was developed by five domain experts. Given that each expert created a different causal graph, the final causal DAG comprises only those edges that were agreed upon by consensus.

\begin{table}[htb!]
\small
    \centering
    \begin{tabular}{l|cc|c}
    \toprule
       \textbf{graph} & \textbf{$\mathbf{\textrm{V}}$} & \textbf{$\mathbf{\textrm{E}}$}  & \textbf{Description} \\ \midrule
        Cancer & $5$ & $4$ & Factors around lung cancer \\
        Survey & $6$ & $6$ & Factors for choosing transportation\\ 
        Asia & $8$ & $8$ & Factors affecting dysponea\\
        Law & $8$ & $20$ & factors around legal system \\
        Alzheimer & $9$ & $16$ & Factors around Alzheimer's Disease \\
        Child & $20$ & $25$ & Lung related illness for a child\\
        Insurance & $27$ & $52$ & Factors affecting car accident insurance \\
        Alarm & $37$ & $46$ & Patient monitoring system\\
    \midrule
    \end{tabular}
    \caption{graph description.}
    \label{tab:data_stats}
\end{table}

\subsection{Reproducibility}
For reproducibility, we used temperature $0$ and top-p value as $1$ across all of the models. We also mentioned the snapshot of the model used. We have also included the prompts and examples below. Our code will be released upon acceptance. The graphs are under CC BY-SA 3.0, which allows us to freely modify the graphs for benchmarking. Our benchmark will be released under the CC BY-SA License. 

GPT-4o, GPT-4 was accessed via API. The rest of the models were run on 1 A100 GPU. Since we used an off-the-shelf LLM, there was no training to be performed. Since many of the models were run by API, it is difficult to calculate the entire computation, however, all of the experiments for each model took $\approx 6$ hours.

\subsection{Controlled Variable Identification}
For variable identification, we generate multiple choices that remain consistent across all missing nodes and all of the graphs. The words were randomly chosen to be far enough from the nodes. The options chosen were weather, book sales, and movie ratings. We wanted to make sure that the options were not from one specific domain, such that the LLM could do the process of elimination. 

\subsection{Causal effect}
\label{app:nde}
\paragraph{Average Treatment Effect.}
Average Treatment Effect (ATE) quantifies the expected change in the outcome \(v_y\) caused by the unit change of the treatment \(v_t\). ATE is a part of the causal do-calculus introduced by \cite{pearl2009causality}. We consider binary causal DAGs, i.e., each variable can either take $0$ or $1$ as values.
\[\text{ATE} = \mathbb{E}[v_y|\text{do}(v_t=1)] - \mathbb{E}[v_y|\text{do}(v_t=0)]\]
where the \(\text{do}(\cdot)\) operator, represents an intervention. The \(E[v_y|\text{do}(v_t=1)]\) represents the expected value of the outcome variable $v_y$ when we intervene to set the treatment variable $v_t$ to $1$ (i.e., apply the treatment), and \(E[v_y|\text{do}(v_t=0)]\) represents the expected value of $v_y$ when we set $v_t$ to $0$ (i.e., do not apply the treatment).

\paragraph{Mediation Analysis.}
Mediation analysis is implemented to quantify the effect of a treatment on the outcome via a third variable, the mediator. The total mediation effect can be decomposed into the Natural Direct Effect (NDE) and the Natural Indirect Effect (NIE). The Natural Direct Effect (NDE) is the effect of the treatment on the outcome variable when not mediated by the mediator variable. The Natural Indirect Effect (NIE) is the effect of the treatment variable on the outcome variable when mediated by the mediator variable.
\begin{align*}
\text{NDE} &= \mathbb{E}[v_{t=1},v_{m=0} - v_{t=0},v_{m=0}]
\end{align*}
Here, NDE is calculated by comparing the expected outcome when the treatment variable is set to $1$ and the mediator is fixed at the level it would take under the control treatment $v_t=0$, with the expected outcome when both the treatment and the mediator are set to the control level.
\begin{align*}
 \text{NIE} &= \mathbb{E}[v_{t=0},v_{m=1} - v_{t=0},v_{m=0}]
\end{align*}
Here, NIE is calculated by comparing the expected outcome when the treatment variable is set to $1$ and the mediator is allowed to change as it would under the treatment, with the expected outcome when the treatment variable is set to $1$ but the mediator is fixed at the control level.

\subsection{Semantic Similarity}
\label{app:semantic}
Given the task of hypothesizing missing nodes in a partial graph \(\mathcal{G}^{*}\) in the absence of  multiple-choices, we evaluate the semantic similarity between the model's predictions and the ground truth node variable. We leverage an open model namely 'all-mpnet-base-v2' to transform the textual representations of the model's predictions and the ground truth into high-dimensional vector space embeddings. Post transforming textual representations into embeddings and normalizing them, we calculate the cosine similarity. Scores closer to 1 indicate a high semantic similarity, suggesting the model's predictions align well with the ground truth. This metric gives a score of similarity without the contextual knowledge of the causal graph. We perform our experiments to consider every node of the ground truth as a missing node iteratively. For all the suggestions for a node variable, we calculate the semantic similarity. The average similarity reported is the highest semantic similarity for each of the variable suggestions.
\begin{algorithm}[h]
\caption{Evaluating Semantic Similarity for Hypothesized Missing Nodes}
\begin{algorithmic}[1]

\State \textbf{Input:} Partial graph $\mathcal{G}^{*}$, Ground truth node variables $V_{\text{GT}}$, Language model $LM = \text{'all-mpnet-base-v2'}$
\State \textbf{Output:} Average highest semantic similarity score

\Procedure{SemanticSimilarity}{$\mathcal{G}^{*}, V_{\text{GT}}, LM$}
    \State Initialize $similarityScores$ as an empty list
    \For{each node $v_{\text{GT}}$ in $\mathbf{v}$}
        \State $predictions \gets$ GeneratePredictions($\mathcal{G}^{*}, LM$)
        \State Initialize $nodeScores$ as an empty list
        \For{each prediction $p$ in $predictions$}
            \State $embedding_{\text{GT}} \gets$ Embed($v_{\text{GT}}, LM$)
            \State $embedding_{p} \gets$ Embed($p, LM$)
            \State Normalize $embedding_{\text{GT}}$ and $embedding_{p}$
            \State $score \gets$ CosineSimilarity($embedding_{\text{GT}}, embedding_{p}$)
            \State Append $score$ to $nodeScores$
        \EndFor
        \State $maxScore \gets$ Max($nodeScores$)
        \State Append $maxScore$ to $similarityScores$
    \EndFor
    \State $averageScore \gets$ Average($similarityScores$)
    \State \textbf{return} $averageScore$
\EndProcedure

\end{algorithmic}
\end{algorithm}

\begin{table}[H]
\tiny
    \centering
    \begin{tabular}{p{3cm} p{2cm} p{2cm} p{2cm} p{2cm} p{2cm}}
    \toprule
    \midrule
        \textbf{Ground Truth}:  &Smoking status  \\  
        \textit{LLM Suggestions}: & Smoking & Alcohol Consumption & Exposure to Radiation & Poor Diet & Genetic Predisposition\\ 
        Semantic similarity : & $0.72$ & $0.38$ &$0.22$ & $0.22$ & $0.17$\\ \midrule

       \textbf{Ground Truth}: & Employee or self-employed  \\  
        \textit{LLM Suggestions}: & Income Level & Job Location & Environmental Awareness & Lifestyle Preferences & Health Consciousness\\ 
        Semantic similarity : & $0.30$ & $0.25$ &$0.17$ & $0.15$ & $0.10$\\ \midrule

       \textbf{Ground Truth}: & Dyspnea laboured breathing  \\  
        \textit{LLM Suggestions}: & Shortness of breath &Chest Pain & Coughing & Fatigue & Weight Loss\\ 
        Semantic similarity : & $0.57$ & $0.41$ &$0.36$ & $0.29$ & $0.11$\\ \midrule

       \textbf{Ground Truth}: & Montreal Cognitive Assessment score  \\  
        \textit{LLM Suggestions}: & Cognitive Function & Neurological Function & Mental Health Status & Risk of Alzheimer's Disease & Memory Performance\\ 
        Semantic similarity : & $0.60$ & $0.47$ &$0.38$ & $0.36$ & $0.16$\\ \midrule

       \textbf{Ground Truth}: & Grunting in infants  \\  
        \textit{LLM Suggestions}: & Respiratory distress & Asthma & Pneumonia & Pulmonary infection & Bronchopulmonary dysplasia (BPD)\\ 
        Semantic similarity : & $0.22$ & $0.18$ &$0.17$ & $0.11$ & $0.01$\\ \midrule

               \textbf{Ground Truth}: & Driving history  \\  
        \textit{LLM Suggestions}: & Previous accidents & Distance driven daily & Type of car insurance & Frequency of car maintenance & Location of parking\\ 
        Semantic similarity : & $0.55$ & $0.42$ &$0.27$ & $0.26$ & $0.18$\\ \midrule

               \textbf{Ground Truth}: & Heart rate blood pressure  \\  
        \textit{LLM Suggestions}: & Pulse Rate & Blood Pressure & Respiratory Rate & EKG Reading & Blood Oxygen Level\\ 
        Semantic similarity : & $0.78$ & $0.78$ &$0.57$ & $0.49$ & $0.42$\\ 

        \\ \midrule

    \bottomrule
    \end{tabular}
    \caption{Examples of model suggestions from and the corresponding semantic similarity score for a missing node variable from each of the graphs.  }
    \label{tab:ssvsllm}
\end{table}

\newpage

\subsection{LLM-as-Judge}
\label{app:sec:llmj}
To capture the domain knowledge of the expert that selects the most relevant causal variable, we use LLM-as-Judge as a proxy expert. This also allows for evaluation based on contextual DAG knowledge as well. Given the impressive results of GPT-4 in~\citep{zheng2023judging}, we use GPT-4 as a judge for all of the experiments. 

\begin{algorithm}[]
\caption{Evaluating Model Suggestions with LLM as Judge}
\begin{algorithmic}[1]

\State \textbf{Input:} Partial graph $\mathcal{G}^{*}$, Ground truth node variables $V_{\text{GT}}$, Predictions $P$, Language model LLM = GPT-4
\State \textbf{Output:} Average quality rating of model's suggestions

\Procedure{LLMasJudge}{$\mathcal{G}^{*}, V_{\text{GT}}, P$, LLM}
    \State Initialize $qualityRatings$ as an empty list
    \For{each node $v_{\text{GT}}$ in $\mathbf{V}$}
        \State $suggestions \gets$ GenerateSuggestions($\mathcal{G}^{*}, P$, LLM)
        \State $bestSuggestion \gets$ SelectBestSuggestion($suggestions, v_{\text{GT}}$, LLM)
        \State $rating \gets$ RateSuggestion($bestSuggestion$, LLM)
        \State Append $rating$ to $qualityRatings$
    \EndFor
    \State $averageRating \gets$ Average($qualityRatings$)
    \State \textbf{return} $averageRating$
\EndProcedure

\Function{GenerateSuggestions}{$\mathcal{G}^{*}, P$, LLM}
    \State \textbf{return} A set of suggestions for missing nodes based on $P$
\EndFunction

\Function{SelectBestSuggestion}{$suggestions, v_{\text{GT}}$,LLM}
    \State Prompt LLM  with $\mathcal{G}^{*}$, $v_{\text{GT}}$, and $suggestions$
    \State \textbf{return} LLM's choice of the best fitting suggestion
\EndFunction

\Function{RateSuggestion}{$suggestion, LM$}
    \State Prompt LLM to rate $suggestion$ on a scale of 1 to 10
    \State \textbf{return} LLM's rating
\EndFunction

\end{algorithmic}
\end{algorithm}

\begin{table}[H]
\small
    \centering
    \begin{tabular}{ll}
    \toprule
    \midrule
        \textbf{Ground Truth}:  &Education up to high school or university degree   \\  
        \textit{Top ranked suggestion}: & Education level\\ 
        Rating : & $9.5$\\ \midrule
        
        \textbf{Ground Truth}:  &Pollution   \\  
        \textit{Top ranked suggestion}: & Smoking history\\ 
        Rating : & $2.0$\\   \midrule  

        \textbf{Ground Truth}:  &Bonchitis   \\  
        \textit{Top ranked suggestion}: & smoking behavior\\ 
        Rating : & $2.0$\\   \midrule 

        \textbf{Ground Truth}:  & Lung XRay report    \\  
        \textit{Top ranked suggestion}: & Lung Damage \\ 
        Rating : & $8.0$\\   \midrule 
        \textbf{Ground Truth}:  & Socioeconomic status    \\  
        \textit{Top ranked suggestion}: &   Driver's lifestyle \\ 
        Rating : & $7.0$\\   \midrule

    \bottomrule
    \end{tabular}
    \caption{Examples of model suggestions from and the corresponding LLM-as-judge score for a missing node variable. }
    \label{tab:ssvsllm2}
\end{table}
\paragraph{Shortcomings of LLM-as-judge.}
LLM-as-judge uses GPT-4 as a judge model which could be biased towards some data. Since the training graphs are not public for this model, it would be hard to judge how these biases might affect the final score. Hence for robust evaluation we also evaluate using the semantic similarity.
\begin{table}[]
\small
    \centering
    \begin{tabular}{lc}
    \toprule
       \textbf{Ground Truth}: Dyspnea laboured breathing  \\  
        \textbf{LLM Suggestion}: Shortness of breath \\ \midrule
        Semantic similarity to GT: $0.57$ \\ 
        LLM-as-Judge score: $9.5$ \\
    \midrule
    \end{tabular}
    \caption{Example comparing the semantic similarity and LLM-as-Judge metrics. Dyspnea is a medical term for shortness of breath. In this example, the contextual information, beyond exact matching, is better captured by LLM-as-Judge. }
    \label{tab:mainssvsllm}
\end{table}

\subsection{Iteratively Hypothesizing in Open World}
For each order, the algorithm prompts the LLM to generate mediator suggestions, selects the suggestion with the highest semantic similarity to the context, and iteratively updates the partial graph with these mediators. $\Delta$, quantifies the impact of mediator ordering by comparing the average highest semantic similarity scores obtained from both descending and ascending orders. This methodical evaluation sheds light on how the sequence in which mediators are considered might affect the LLM's ability to generate contextually relevant and accurate predictions.

\begin{algorithm}
\caption{ Random Order Mediator Hypothesis}
\begin{algorithmic}[1]
\State \textbf{Input:} Partial graph $\mathcal{G}^*$ (where $\mathcal{G}^* = \mathcal{G} - H$), Treatment $v_t$, Outcome $v_y$, Number of mediators $H$, Number of suggestions $k$
\State \textbf{Output:} Updated graph $\mathcal{G}^*$ with selected mediators

\Procedure{GenerateMediatorsRandom}{$\mathcal{G}^*, v_t, v_y, H, k$}
    \For{$i \gets 1$ to $H$}
        \State $suggestions \gets$ Generate $k$ suggestions for $v_{m_i}$ using $P_{\text{LLM}}(\mathcal{G}^*)$
        \State Initialize $highestSimilarity \gets 0$
        \State Initialize $selectedMediator \gets \text{null}$
        \For{each $suggestion$ in $suggestions$}
            \State $similarityScore \gets$ Calculate semantic similarity for $suggestion$
            \If{$similarityScore > highestSimilarity$}
                \State $highestSimilarity \gets similarityScore$
                \State $selectedMediator \gets suggestion$
            \EndIf
        \EndFor
        \State Update $\mathcal{G}^* \gets \mathcal{G}^* \cup \{selectedMediator\}$
    \EndFor
    \State \textbf{return} $\mathcal{G}^*$
\EndProcedure
\end{algorithmic}
\end{algorithm}

\begin{algorithm}
\caption{Ordered Mediator Generation and Evaluation Based on MIS}
\begin{algorithmic}[1]
\State \textbf{Input:} Partial graph $\mathcal{G}^*$, Treatment $v_t$, Outcome $v_y$, Set of potential mediators $M$, Number of suggestions $k$
\State \textbf{Output:} $\Delta$ - measure of the influence of mediator ordering

\Procedure{CalculateMIS}{$v_t, v_y, M$}
    \State Initialize $\text{MISList}$ as an empty list
    \For{each mediator $v_{m_i}$ in $M$}
        \State Calculate $\text{NIE}(v_{m_i})$ and $\text{NDE}(v_{m_i})$
        \State $\text{MIS}(v_{m_i}) \gets \frac{\text{NIE}(v_{m_i})}{\text{NDE}(v_{m_i})}$
        \State Append $\text{MIS}(v_{m_i})$ to $\text{MISList}$
    \EndFor
    \State \textbf{return} $\text{MISList}$
\EndProcedure

\Procedure{GenerateMediatorsOrdered}{$\mathcal{G}^*, v_t, v_y, M, k$}
    \State $\text{MISList} \gets$ \Call{CalculateMIS}{$v_t, v_y, M$}
    \State Sort $M$ in descending order of $\text{MISList}$ to get $M_{\text{desc}}$
    \State Sort $M$ in ascending order of $\text{MISList}$ to get $M_{\text{asc}}$
    \State $averageDesc \gets$ \Call{GenerateAndEvaluate}{$\mathcal{G}^*, M_{\text{desc}}, k$}
    \State $averageAsc \gets$ \Call{GenerateAndEvaluate}{$\mathcal{G}^*, M_{\text{asc}}, k$}
    \State $\Delta \gets \frac{|averageDesc - averageAsc|}{averageDesc}$
    \State \textbf{return} $\Delta$
\EndProcedure

\Function{GenerateAndEvaluate}{$\mathcal{G}^*, M_{\text{order}}, k$}
    \State Initialize $\text{similarityScores}$ as an empty list
    \For{each mediator $v_{m_i}$ in $M_{\text{order}}$}
        \State Perform the same steps as in the refined random order mediator generation
        \State (Generate $k$ suggestions, select the most similar, update $\mathcal{G}^*$)
        \State Append the highest similarity score to $\text{similarityScores}$
    \EndFor
    \State \textbf{return} Average of $\text{similarityScores}$
\EndFunction

\end{algorithmic}
\end{algorithm}

\newpage
\section{Further results}
\label{app:sec:nde}

\subsection{Variances}
\label{app:tab:open_world}

For brevity we didnt add variance in the main text, the following results have variances:
\begin{table}[H]
\small
\setlength{\tabcolsep}{1.5pt}
    \centering
    \begin{tabular}{l|cc|cc|cc|cc|cc|cc|cc|cc}
    \toprule
        & \multicolumn{2}{c|}{{Cancer}} & \multicolumn{2}{c|}{{Survey}} & \multicolumn{2}{c|}{{Asia}} & \multicolumn{2}{c|}{{Alzheimers}} & \multicolumn{2}{c|}{{Child}} & \multicolumn{2}{c|}{{Insurance}} & \multicolumn{2}{c|}{{Alarm}}   & \multicolumn{2}{c}{{Avg}} \\  \cmidrule{2-17}
      & \tiny{Sim} & \tiny{LLM-J} & \tiny{Sim} & \tiny{LLM-J} & \tiny{Sim} & \tiny{LLM-J} & \tiny{Sim} & \tiny{LLM-J} & \tiny{Sim} & \tiny{LLM-J} & \tiny{Sim} & \tiny{LLM-J} & \tiny{Sim} & \tiny{LLM-J} & \tiny{Sim} & \tiny{LLM-J}\\ \midrule

Zephyr & $\underset{\pm{0.04}}{0.36}$ & $\underset{\pm{0.06}}{0.61}$ & $\underset{\pm{0.07}}{0.34}$ & $\underset{\pm{0.05}}{0.60}$ & $\underset{\pm{0.05}}{0.45}$ & $\underset{\pm{0.04}}{0.66}$ & $\underset{\pm{0.03}}{0.35}$ & $\underset{\pm{0.03}}{0.75}$ & $\underset{\pm{0.02}}{0.51}$ & $\underset{\pm{0.04}}{0.70}$ & $\underset{\pm{0.04}}{0.45}$ & $\underset{\pm{0.05}}{0.44}$ & $\underset{\pm{0.03}}{0.46}$ & $\underset{\pm{0.02}}{0.69}$ & $\underset{\pm{0.04}}{0.42}$ & $\underset{\pm{0.04}}{0.63}$ \vspace{1mm}\\ 
Mixtral & $\underset{\pm{0.03}}{0.41}$ & $\underset{\pm{0.04}}{0.66}$ & $\underset{\pm{0.05}}{0.39}$ & $\underset{\pm{0.06}}{0.66}$ & $\underset{\pm{0.02}}{\mathbf{0.66}}$ & $\underset{\pm{0.03}}{0.75}$ & $\underset{\pm{0.04}}{0.31}$ & $\underset{\pm{0.02}}{0.77}$ & $\underset{\pm{0.03}}{\mathbf{0.53}}$ & $\underset{\pm{0.02}}{\mathbf{0.77}}$ & $\underset{\pm{0.03}}{0.46}$ & $\underset{\pm{0.04}}{\mathbf{0.56}}$ & $\underset{\pm{0.03}}{\mathbf{0.50}}$ & $\underset{\pm{0.06}}{0.72}$ & $\underset{\pm{0.03}}{0.46}$  & $\underset{\pm{0.05}}{0.70}$ \vspace{1mm}\\

Neural & $\underset{\pm{0.02}}{0.38}$ & $\underset{\pm{0.05}}{0.77}$ & $\underset{\pm{0.02}}{0.43}$ & $\underset{\pm{0.03}}{0.55}$ & $\underset{\pm{0.03}}{0.53}$ & $\underset{\pm{0.04}}{0.55}$ & $\underset{\pm{0.05}}{0.44}$ & $\underset{\pm{0.03}}{0.71}$ & $\underset{\pm{0.04}}{0.48}$ & $\underset{\pm{0.03}}{0.70}$ & $\underset{\pm{0.04}}{0.47}$ & $\underset{\pm{0.05}}{0.43}$ & $\underset{\pm{0.02}}{0.47}$ & $\underset{\pm{0.03}}{0.67}$ & $\underset{\pm{0.03}}{0.45}$ & $\underset{\pm{0.04}}{0.63}$\vspace{1mm}\\
        
Llama & $\underset{\pm{0.03}}{0.40}$ & $\underset{\pm{0.05}}{0.48}$ & $\underset{\pm{0.04}}{0.40}$ & $\underset{\pm{0.05}}{0.54}$ & $\underset{\pm{0.03}}{0.53}$ & $\underset{\pm{0.06}}{0.58}$ & $\underset{\pm{0.05}}{0.45}$ & $\underset{\pm{0.03}}{0.61}$ & $\underset{\pm{0.04}}{0.48}$ & $\underset{\pm{0.03}}{0.63}$ & $\underset{\pm{0.01}}{0.42}$ & $\underset{\pm{0.05}}{0.34}$ & $\underset{\pm{0.02}}{0.46}$ & $\underset{\pm{0.03}}{0.65}$ & $\underset{\pm{0.03}}{0.45}$  & $\underset{\pm{0.04}}{0.55}$ \vspace{1mm}\\

Mistral & $\underset{\pm{0.01}}{0.33}$ & $\underset{\pm{0.05}}{0.67}$ & $\underset{\tiny{\pm{0.05}}}{0.44}$ & $\underset{\pm{0.04}}{0.65}$ & $\underset{\pm{0.03}}{0.60}$ & $\underset{\pm{0.04}}{0.73}$ & $\underset{\pm{0.04}}{0.34}$ & $\underset{\pm{0.02}}{0.76}$ & $\underset{\pm{0.04}}{0.48}$ & $\underset{\pm{0.03}}{0.68}$ & $\underset{\pm{0.03}}{0.46}$ & $\underset{\pm{0.01}}{0.47}$ & $\underset{\pm{0.03}}{0.47}$ & $\underset{\pm{0.03}}{0.71}$ & $\underset{\pm{0.03}}{0.44}$ & $\underset{\pm{0.03}}{0.67}$ \vspace{1mm}\\

        
GPT-4 & $\underset{\pm{0.02}}{\mathbf{0.49}}$ & $\underset{\pm{0.03}}{\mathbf{0.90}}$ & $\underset{\pm{0.06}}{\mathbf{0.51}}$ & $\underset{\pm{0.04}}{0.67}$ & $\underset{\pm{0.02}}{\mathbf{0.66}}$ & $\underset{\pm{0.03}}{\mathbf{0.76}}$ & $\underset{\pm{0.02}}{\mathbf{0.47}}$ & $\underset{\pm{0.02}}{0.98}$ & $\underset{\pm{0.05}}{0.36}$ & $\underset{\pm{0.04}}{0.53}$ & $\underset{\pm{0.03}}{\mathbf{0.52}}$ & $\underset{\pm{0.03}}{\mathbf{0.56}}$ & $\underset{\pm{0.06}}{0.49}$ & $\underset{\pm{0.02}}{\mathbf{0.75}}$ & $\underset{\pm{0.04}}{\mathbf{0.50}}$ & $\underset{\pm{0.03}}{\mathbf{0.73}}$ \vspace{1mm}\\

    \bottomrule
    \end{tabular}
    \caption{Average semantic similarity and LLM-as-Judge metrics to evaluate LLMs in hypothesizing the missing variable in a causal DAG.}
\end{table}

\begin{table}[H]
\vspace{-1mm}
\small
    \centering
    \begin{tabular}{l|cc|cc|cc|cc}
    \toprule
   & \multicolumn{2}{c|}{Asia} & \multicolumn{2}{c|}{Child}  & \multicolumn{2}{c|}{Insurance} & \multicolumn{2}{c}{Alarm} \\ 
   & Sim & $\Delta$ & Sim & $\Delta$ & Sim & $\Delta$ & Sim & $\Delta$ \\ \midrule
    Zephyr & $\underset{\pm0.03}{0.61}$ & $\underset{\pm0.01}{-0.02}$ & $\underset{\pm0.04}{\mathbf{0.54}}$  & $\underset{\pm0.02}{0.17}$ & $\underset{\pm0.05}{0.47}$ & $\underset{\pm0.02}{0.19}$ & $\underset{\pm0.05}{0.51}$ & $\underset{\pm0.02}{0.20}$ \\
    Mixtral & $\underset{\pm0.02}{\mathbf{0.87}}$ & $\underset{\pm0.01}{0.01}$ & $\underset{\pm0.05}{0.50}$ &  $\underset{\pm0.02}{0.18}$ & $\underset{\pm0.05}{0.48}$ & $\underset{\pm0.02}{0.15}$ & $\underset{\pm0.05}{0.52}$ & $\underset{\pm0.01}{0.13}$\\
    Neural & $\underset{\pm0.06}{0.65}$ & $\underset{\pm0.02}{0.04}$ & $\underset{\pm0.05}{0.48}$ &  $\underset{\pm0.02}{0.21}$ & $\underset{\pm0.04}{0.42}$ & $\underset{\pm0.02}{0.16}$ & $\underset{\pm0.04}{0.46}$ & $\underset{\pm0.01}{0.12}$\\
    Llama & $\underset{\pm0.08}{0.80}$ & $\underset{\pm0.02}{0.07}$ & $\underset{\pm0.05}{0.49}$ &  $\underset{\pm0.01}{-0.05}$ & $\underset{\pm0.06}{0.44}$ & $\underset{\pm0.02}{0.21}$ & $\underset{\pm0.05}{0.51}$  & $\underset{\pm0.01}{0.07}$\\
    Mistral & $\underset{\pm0.03}{0.33}$ & $\underset{\pm0.01}{0.02}$ & $\underset{\pm0.05}{0.50}$ & $\underset{\pm0.01}{0.12}$ & $\underset{\pm0.05}{0.48}$ & $\underset{\pm0.02}{0.13}$ & $\underset{\pm0.04}{0.47}$ & $\underset{\pm0.01}{0.11}$\\
    GPT-4 & $\underset{\pm0.07}{0.49}$ & $\underset{\pm0.01}{0.04}$ & $\underset{\pm0.05}{0.39}$ & $\underset{\pm0.02}{0.16}$ & $\underset{\pm0.05}{\mathbf{0.52}}$ & $\underset{\pm0.02}{0.14}$ & $\underset{\pm0.06}{\mathbf{0.60}}$ & $\underset{\pm0.01}{-0.07}$\\    \midrule
    \end{tabular}
    \caption{Sim: semantic similarity for iteratively hypothesizing the mediator nodes when prompted with random order. $\Delta$ measures the change in the prediction of each model according to the MIS. }
    \label{tab:app:iterM}
\end{table}



\subsection{Breaking down the performance}
\begin{figure}[H]
\centering
    \begin{subfigure}[t]{0.35\textwidth}
         \centering
         \includegraphics[width=\textwidth]{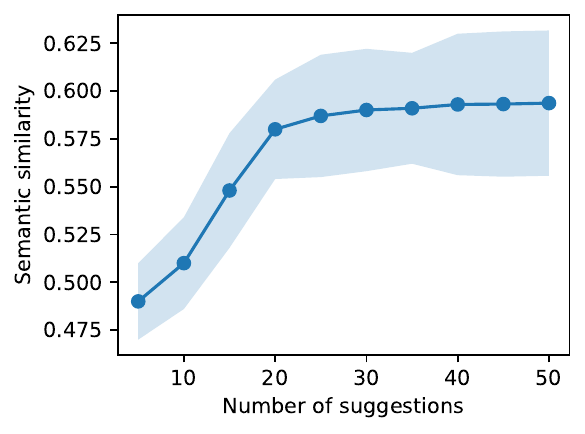}
          \caption{}
         \label{fig:topk}
     \end{subfigure}
    \begin{subfigure}[t]{0.35\textwidth}
         \centering
         \includegraphics[width=\textwidth]{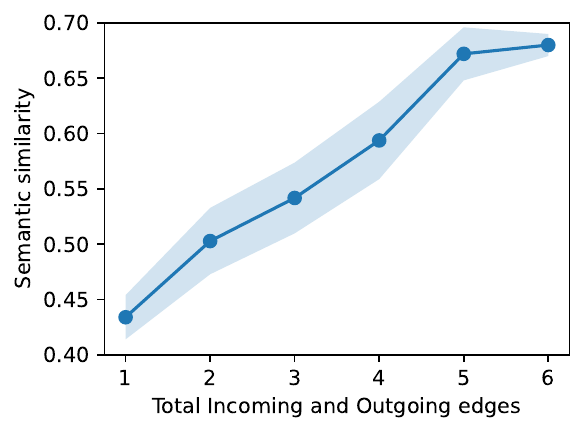}
         \caption{}
         \label{fig:nodevsacc}
     \end{subfigure}
     \vspace{-2mm}
     \caption{L: Plot of semantic similarity with an increasing number of suggestions for GPT-4 on the Alarm graph. R: Plot of semantic similarity against the total number of incoming and outgoing edges for GPT-4 on the Alarm graph. }
     \vspace{-3mm}
\end{figure}

\subsection{Effect of context}
We observed notable differences in the accuracy of LLM predictions for missing nodes within causal graphs when context was provided versus when it was absent. Specifically, the inclusion of contextual information about the causal graph significantly enhanced the LMs' ability to generate accurate and relevant predictions. In realistic settings, when this setup is being used by a scientist, they would provide the context of the task along with the partial graph. When context was not provided, the models often struggled to identify the most appropriate variables, leading to a decrease in prediction accuracy, especially for smaller models. Unsurprisingly, providing context was more important for smaller graphs than larger graphs. LLMs were able to understand the context of the graph via multiple other nodes in the graph for larger graphs. 

\begin{table*}[htb!]
\small
\setlength{\tabcolsep}{2pt}
    \centering
    \begin{tabular}{l|cc|cc|cc|cc|cc}
    \toprule
        & \multicolumn{2}{c|}{{Cancer}} & \multicolumn{2}{c|}{{Survey}} & \multicolumn{2}{c|}{{Asia}} &\multicolumn{2}{c|}{{Insurance}}  & \multicolumn{2}{c}{{Alarm}}\\  
        & $X$ & $\checkmark$ & $X$ & $\checkmark$ & $X$ & $\checkmark$ & $X$ & $\checkmark$ & $X$ & $\checkmark$ \\  \midrule

        In-Context & $0.75$ & $ 1.00$ & $0.67$ & $1.00$ & $ 0.68$& $ 0.88$ & $0.85$&$0.90$ &$0.96$& $0.96 $ \\
        Out-of-Context & $0.00$ & $ 0.25$ & $0.33$ & $0.33$ & $ 0.53$ & $0.61$ & $ 0.58$ & $0.58$ & $0.60 $ & $0.57$ \\
        Open world Hypothesis & $0.39$ & $0.41$ & $0.40$ & $ 0.39$ & ${0.63}$ & $ 0.66$ & $ 0.49$ & $0.50 $ & $0.44$ & $0.46$ \\

    \bottomrule
    \end{tabular}
    \caption{Model-Mixtral to evaluate the effect of context given in the prompt.}
    \label{app:context}
\end{table*}


\subsection{Using explanations}
\label{app:llmjudgevs}
While using LLMs for hypothesizing the missing nodes within the causal graph for the open world setting, an additional question is for the model to provide explanations for each of its predictions. This was motivated by the fact that incorporating a rationale behind each prediction might enhance the model's semantic similarity. We present the results in the Table below. We observe that evaluating semantic similarity with explanations leads to a decrease in performance as compared to the earlier setting where the language model returned phrases. This is because semantic similarity, as a metric, evaluates the closeness of the model's predictions to the ground truth in a high-dimensional vector space, focusing on the semantic content encapsulated within the embeddings. It is a metric that leaves little room for interpretative flexibility, focusing strictly on the degree of semantic congruence between the predicted and actual variables. The introduction of explanations, while enriching the model's outputs with contextual insights, did not translate into improved semantic alignment with the ground truth.

\begin{table*}[htb!]
\small
\setlength{\tabcolsep}{2pt}
    \centering
    \begin{tabular}{l|cc|cc|cc|cc|cc}
    \toprule
        & \multicolumn{2}{c|}{{Cancer}} & \multicolumn{2}{c|}{{Survey}} & \multicolumn{2}{c|}{{Asia}} &\multicolumn{2}{c|}{{Insurance}}  & \multicolumn{2}{c}{{Alarm}}\\  
        & $X$ & $\checkmark$ & $X$ & $\checkmark$ & $X$ & $\checkmark$ & $X$ & $\checkmark$ & $X$ & $\checkmark$ \\  \midrule

        Sim & $\underset{\pm{0.02}}{0.49}$ & $\underset{\pm{0.07}}{0.38}$ & $\underset{\pm{0.06}}{0.51}$ & $\underset{\pm{0.10}}{0.44}$ &  $\underset{\pm{0.02}}{0.66}$ & $\underset{\pm{0.09}}{0.57}$ &  $\underset{\pm{0.03}}{0.52}$ & $\underset{\pm{0.07}}{0.40}$ & $\underset{\pm{0.06}}{0.49}$& $\underset{\pm{0.06}}{0.40}$\\
        
        LLM-Judge &  $\underset{\pm{0.03}}{0.90}$ & $\underset{\pm{0.02}}{0.91}$ & $\underset{\pm{0.04}}{0.67}$ & $\underset{\pm{0.02}}{0.69}$ &  $\underset{\pm{0.03}}{0.76}$ & $\underset{\pm{0.04}}{0.76}$ &  $\underset{\pm{0.03}}{0.56}$ & $\underset{\pm{0.03}}{0.55}$ & $\underset{\pm{0.02}}{0.75}$& $\underset{\pm{0.02}}{0.75}$\\
    \bottomrule
    \end{tabular}
    \caption{Model-GPT 4. Evaluating the effect of explanations on different metrics from Task 3.}
    \label{app:expl}
\end{table*}
 
\paragraph{Ambiguous predictions which semantically represent the same variable. }
An important linguistic concern that could be missed by semantic similarity is an ambiguous hypothesis by the LLM that may have the same semantics, which again breaks the semantic similarity metric. This further motivates the LLM-judge metric, whose input is the context of the causal graph, the partial causal graph, the ground truth variable, and the model predictions. Given the rich context of the LLM-judge metric, we suspect it would be able to overcome the ambiguity. We prompted the model to justify its hypothesis variables using explanations. We observe that evaluating semantic similarity with explanations leads to a decrease in performance as compared to the earlier setting where the language model returned just phrases. In Table~\ref {app:expl}, we observed a drop in performance for semantic similarity. In contrast, we observe a similar or slight improvement in the LLM-judge metric when the explanation of the model hypothesis is given. 

\subsection{Chain of thought}
Chain-of-Thought prompting has gained popularity due to its impressive performance in proving the quality of LLMs' output~\cite{kojima2022large}, also in metadata-based causal reasoning~\cite{vashishtha2023causal}. We also incorporated COT prompting for our prompts. We perform ablation studies in \autoref{app:cot}. We observe that COT particularly improves the performance of the identification experiments. 
\begin{table*}[htb!]
\small
\setlength{\tabcolsep}{2pt}
    \centering
    \begin{tabular}{l|cc|cc|cc|cc|cc}
    \toprule
        & \multicolumn{2}{c|}{{Cancer}} & \multicolumn{2}{c|}{{Survey}} & \multicolumn{2}{c|}{{Asia}} &\multicolumn{2}{c|}{{Insurance}}  & \multicolumn{2}{c}{{Alarm}}\\  
        & $X$ & $\checkmark$ & $X$ & $\checkmark$ & $X$ & $\checkmark$ & $X$ & $\checkmark$ & $X$ & $\checkmark$ \\  \midrule

        In-Context & $1.00$ & $ 1.00$ & $0.83$ & $1.00$ & $ 0.75$& $ 0.88$ & $0.74$&$0.90$ &$0.91$& $0.96 $ \\
        Out-of-Context & $0.50$ & $ 0.25$ & $0.18$ & $0.33$ & $ 0.57$ & $0.61$ & $ 0.56$ & $0.58$ & $0.54 $ & $0.57$ \\

    \bottomrule
    \end{tabular}
    \caption{Model-Mixtral to evaluate the effect of COT given in the prompt.}
    \label{app:cot}
\end{table*}

\subsection{Iterative mediator search vs all at once}
\label{app:iterativevsall}
For Task 4, we iteratively hypothesize the missing variables (mediators). Our choice was primarily driven by the complexity of Task 4, which involves predicting multiple missing mediators, ranging from 1 to 10. For a Task with 10 missing mediators, the model would have to predict 50 suggestions at once. We initially hypothesized that LLMs might struggle with making multiple predictions across different variables simultaneously. This was indeed reflected in our results and GPT-4 outputs from Table X. The iterative approach allows the model's prediction to narrow the search space, which would not be possible in a non-iterative approach. This method is more aligned with the scientific discovery process, where hypotheses are often refined iteratively based on new findings.
Furthermore, our approach simulates a human-in-the-loop scenario, where the most plausible answer is selected and used to guide the next prediction.

\begin{table}[!ht]
    \centering
    \begin{tabular}{l|c|c|c|c}
    \midrule
        ~ & Asia & Child & Insurance & Alarm \\ \midrule
        Non-iterative & 0.42 +- 0.07 & 0.33 +- 0.06 & 0.45 +- 0.09 & 0.54 +- 0.05 \\ 
        Iterative & 0.49 +- 0.05 & 0.39 +-0.03 & 0.52 +- 0.02 & 0.60 +- 0.04 \\ \midrule
    \end{tabular}
\end{table}

{\subsection{Results on Neuropathic graph}
We added a new graph, the neuropathic pain graph~\cite{tu2019neuropathic}, which is not part of common LLM training corpora as one needs to use a python script to download it.  The graph consists of 221 nodes and 770 edges, but for feasibility, we selected a subset of the graph for evaluation. We ran experiments for Task 1, Task 2, and Task 3.
\begin{table}[h!]
\centering
\begin{tabular}{l|c|cc|cc}
\toprule
{Model} & {Task 1} & {Task 2 Result} & {Task 2 FNA} & {Task 3 Sim} & {Task 3 LLM-J} \\
\midrule
{Mistral} & {0.64} & {0.51} & {0.32} & {0.38} & {0.53} \\
{Mixtral} & {0.83} & {0.55} & {0.34} & {0.45} & {0.69} \\
{Llama}   & {0.78} & {0.49} & {0.27} & {0.44} & {0.63} \\
{GPT-4}   & {0.94} & {0.68} & {0.24} & {0.51} & {0.76} \\
\bottomrule
\end{tabular}
\caption{Comparison of model performances across tasks on Neuropathic graph.}
\label{tab:task_performance}
\end{table}
}

\subsection{Fine grained model performance}
\begin{figure}[H]
\centering
     \begin{subfigure}[t]{0.24\textwidth}
         \centering
         \includegraphics[width=\textwidth]{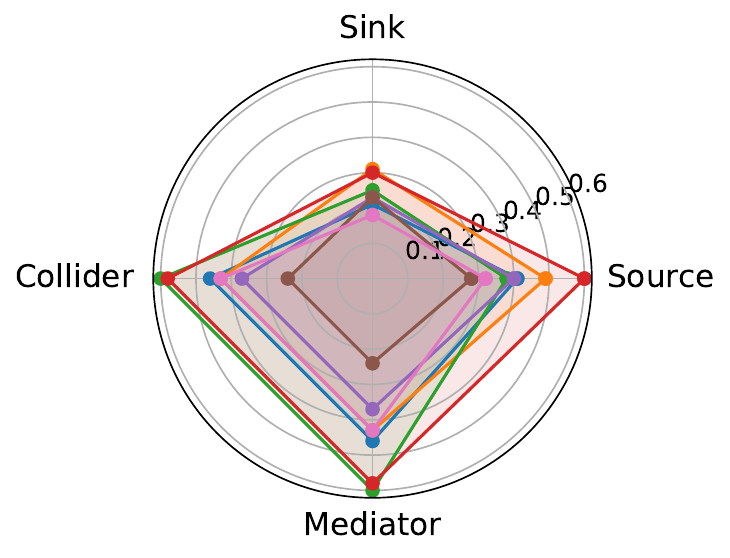}
         \caption{Cancer}
     \end{subfigure}
     \begin{subfigure}[t]{0.24\textwidth}
         \centering
         \includegraphics[width=\textwidth]{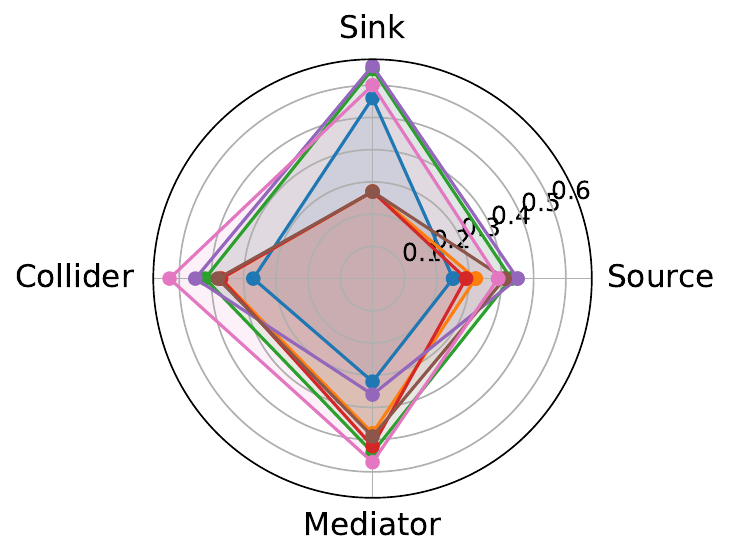}
         \caption{Survey}
     \end{subfigure}
    \begin{subfigure}[t]{0.24\textwidth}
         \centering
         \includegraphics[width=\textwidth]{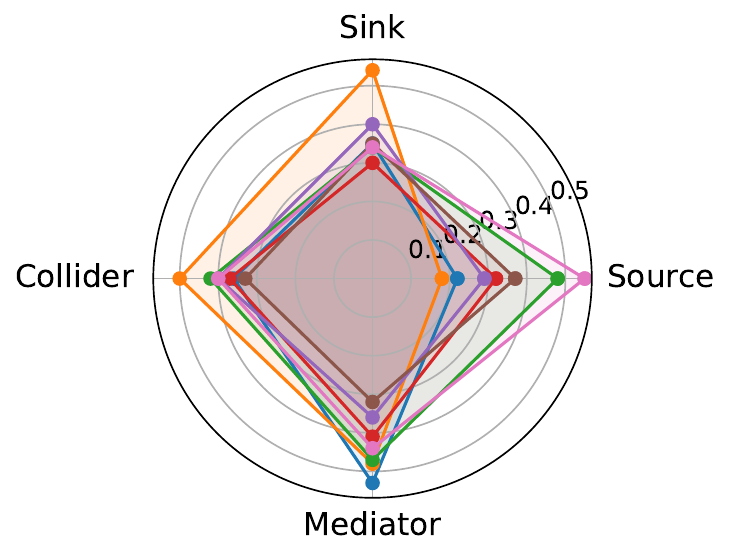}
         \caption{Alzheimers}
     \end{subfigure}
     \begin{subfigure}[t]{0.25\textwidth}
         \centering
         \includegraphics[width=\textwidth]{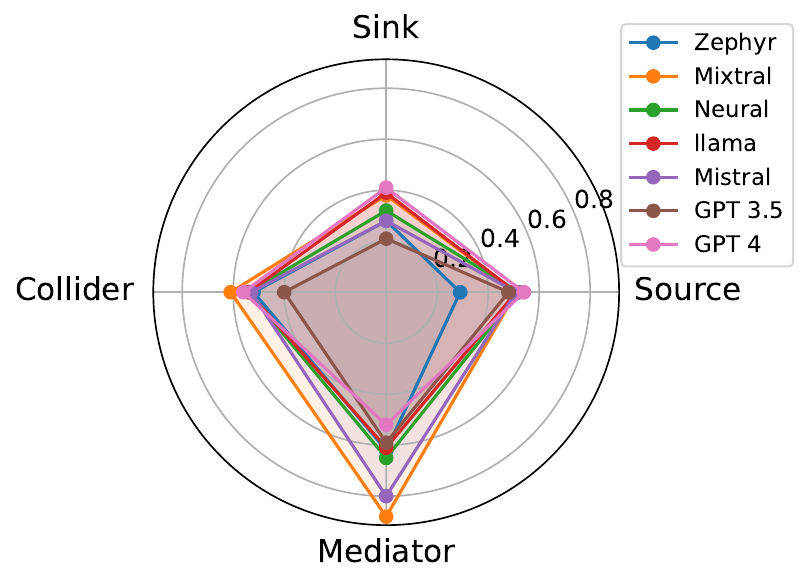}
         \caption{Asia}
     \end{subfigure}
    \begin{subfigure}[t]{0.24\textwidth}
         \centering
         \includegraphics[width=\textwidth]{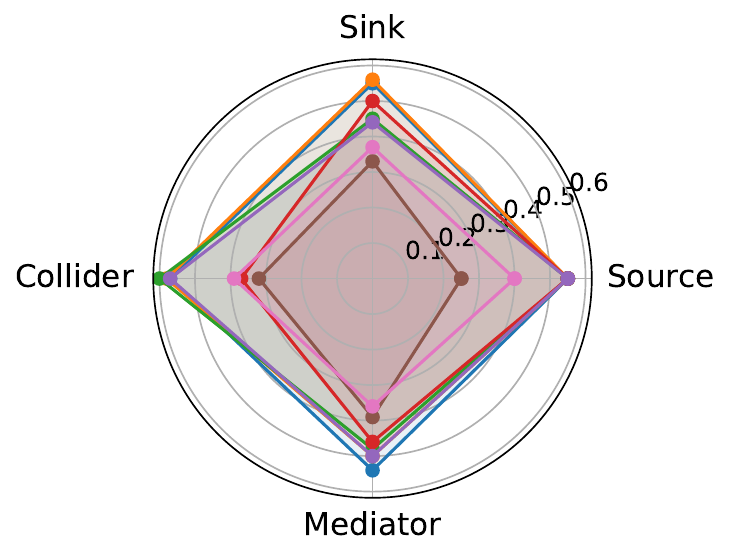}
         \caption{Child}
     \end{subfigure}
    \begin{subfigure}[t]{0.24\textwidth}
         \centering
         \includegraphics[width=\textwidth]{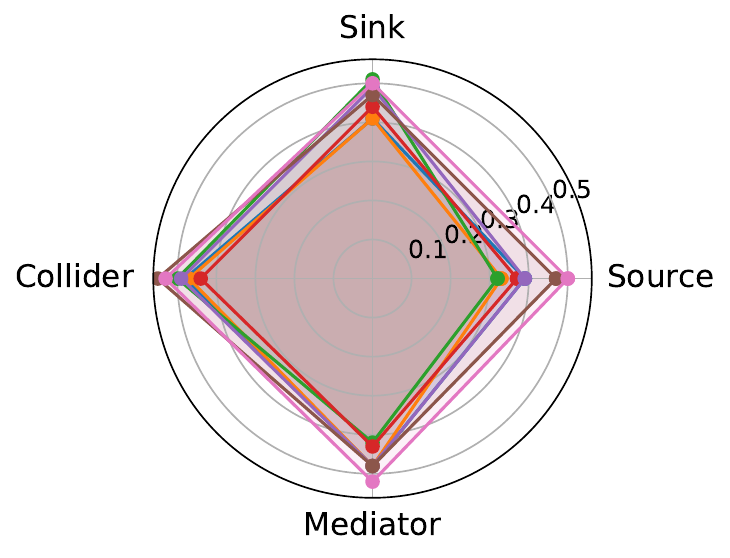}
         \caption{Insurance}
     \end{subfigure}
    \begin{subfigure}[t]{0.24\textwidth}
         \centering
         \includegraphics[width=\textwidth]{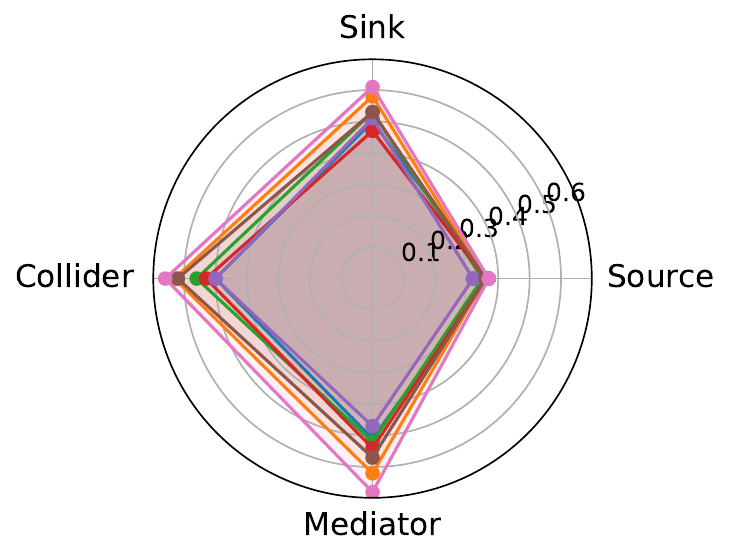}
         \caption{Alarm}
     \end{subfigure}
     \caption{Detailed spider plots for Semantic similarity}
     \label{fig:spider_detail}
\end{figure}

\begin{figure} [H]
\centering
     \begin{subfigure}[t]{0.24\textwidth}
         \centering
         \includegraphics[width=\textwidth]{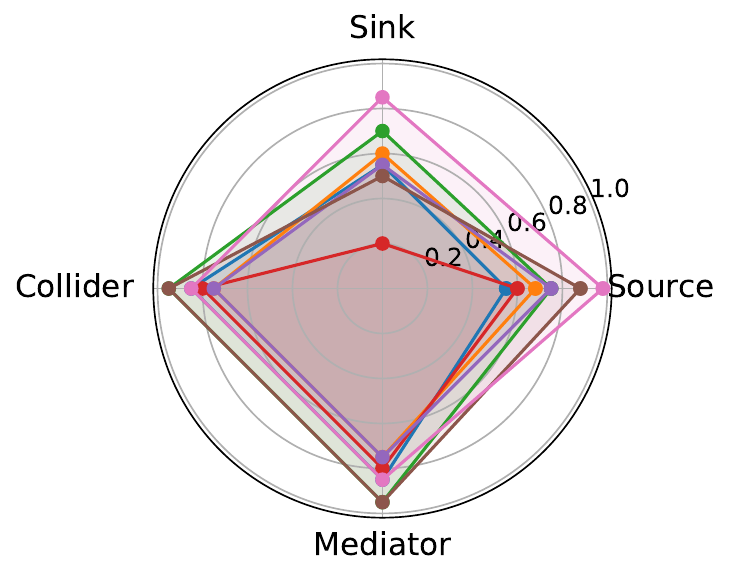}
         \caption{Cancer}
     \end{subfigure}
     \begin{subfigure}[t]{0.24\textwidth}
         \centering
         \includegraphics[width=\textwidth]{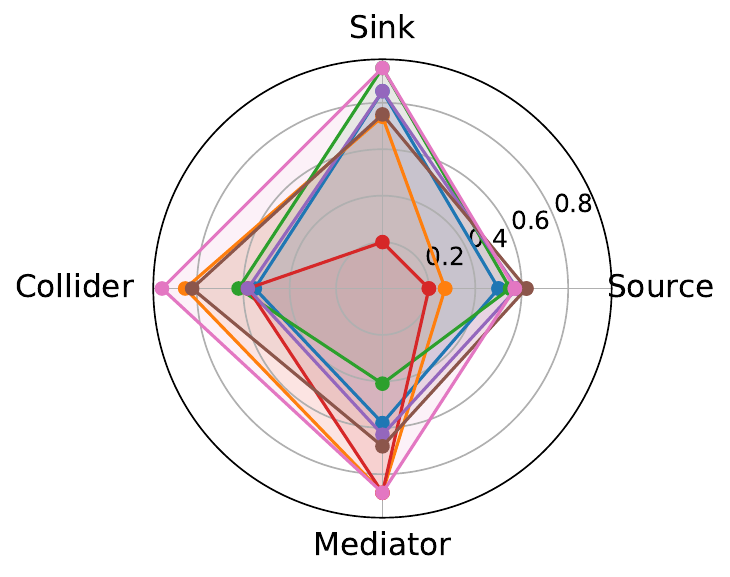}
         \caption{Survey}
     \end{subfigure}
    \begin{subfigure}[t]{0.24\textwidth}
         \centering
         \includegraphics[width=\textwidth]{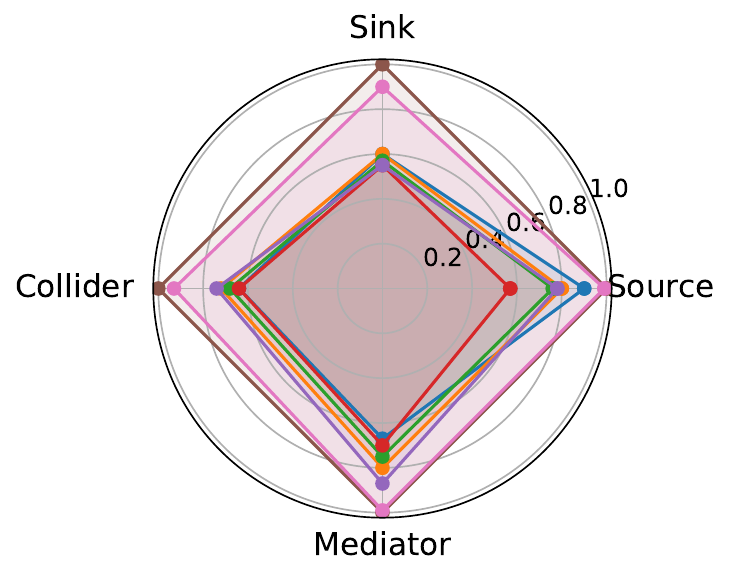}
         \caption{Alzheimers}
     \end{subfigure}
     \begin{subfigure}[t]{0.24\textwidth}
         \centering
         \includegraphics[width=\textwidth]{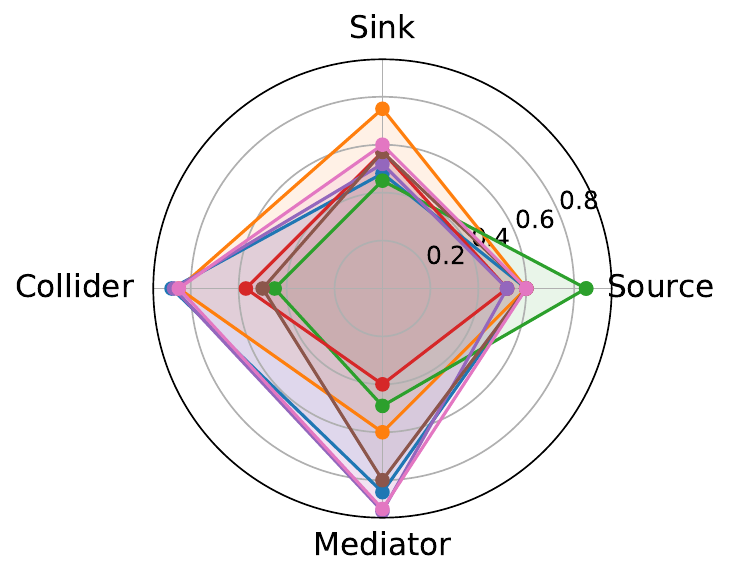}
         \caption{Asia}
     \end{subfigure}
    \begin{subfigure}[t]{0.24\textwidth}
         \centering
         \includegraphics[width=\textwidth]{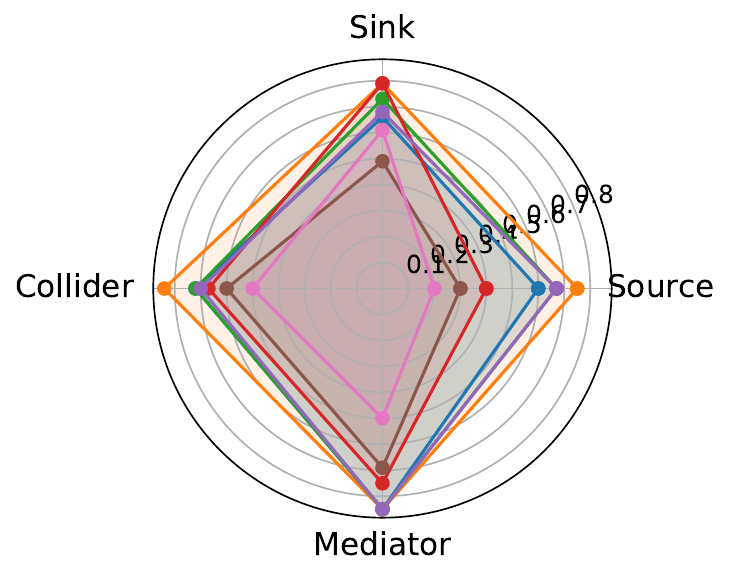}
         \caption{Child}
     \end{subfigure}
    \begin{subfigure}[t]{0.24\textwidth}
         \centering
         \includegraphics[width=\textwidth]{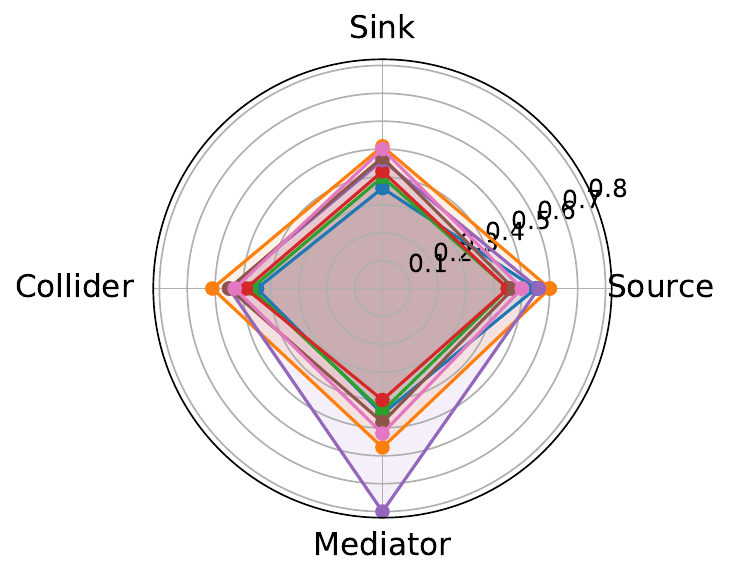}
         \caption{Insurance}
     \end{subfigure}
    \begin{subfigure}[t]{0.24\textwidth}
         \centering
         \includegraphics[width=\textwidth]{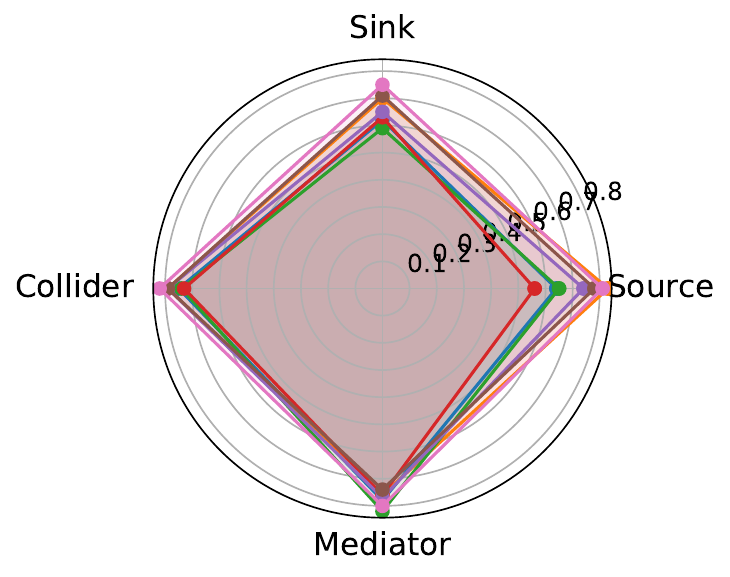}
         \caption{Alarm}
     \end{subfigure}
     \caption{Detailed spider plots for LLM-as-judge metric}
     \label{fig:spider_detail}
\end{figure}

\subsection{Testing whether LLM is using context or parametric graphs for GPT*}

\begin{table}[H]
\centering
\begin{tabular}{lccc}
\toprule
\textbf{Dataset} & \textbf{Current Sim} & \textbf{Memorization} & \textbf{Est. Sim w/ BNLearn Context} \\
\midrule
Cancer       & 0.49 & \checkmark & 0.60 \\
Survey       & 0.51 & \checkmark & 0.62 \\
Asia         & 0.66 & \checkmark & 0.78 \\
Law          & 0.55 & \ding{55}   & 0.56 \\
Alzheimers   & 0.47 & 0.55        & 0.52 \\
Child        & 0.36 & \checkmark & 0.52 \\
Insurance    & 0.52 & \checkmark & 0.64 \\
Alarm        & 0.49 & \checkmark & 0.62 \\

\bottomrule
\end{tabular}
\caption{Estimated similarity improvement for GPT-4 when informed that graphs are from the BNLearn repository. The memorization column shows whether GPT-4 recalled structural details.}
\label{tab:bnlearn_context_effect}
\end{table}

To test whether GPT-4’s original performance was driven by the retrieval of memorized content, we reran the variable inference task with explicit prompts stating that the graphs were from the BNLearn repository. We observed modest gains in similarity for well-known graphs (e.g., Asia, Alarm), indicating that GPT-4 can retrieve additional details when cued. However, the performance in the original setting, without such cues, was already strong, suggesting that the model was not merely retrieving memorized structures. Instead, its responses appear to reflect contextual reasoning and generalization beyond rote recall.

\subsection{Converting causal graph to prompt}
\label{app:encoding}
We observe that different graph representations yield similar performance across tasks, with the most variation for Task 2 where we have 2 missing variables on Mistral and Mixtral models. 
\begin{table}[H]
\centering
\begin{tabular}{lcccc}
\hline
Model          & Asia & Child & Insurance & Alarm \\
\hline
JSON           & 0.80 & 0.79  & 0.50      & 0.85  \\
GraphML        & 0.80 & 0.78  & 0.47      & 0.85  \\
Textual (ours) & 0.78 & 0.80  & 0.49      & 0.85  \\
\hline
\end{tabular}
\caption{Different encoding strategies for Task 1}
\end{table}

\vspace{1em}

\begin{table}[H]
\centering
\begin{tabular}{lcccc}
\hline
Model          & Asia      & Child     & Insurance & Alarm     \\
\hline
JSON           & 0.73/0.16 & 0.45/0.30 & 0.37/0.17 & 0.50/0.21 \\
GraphML        & 0.70/0.15 & 0.41/0.29 & 0.37/0.12 & 0.53/0.22 \\
Textual (ours) & 0.73/0.17 & 0.42/0.31 & 0.40/0.12 & 0.51/0.22 \\
\hline
\end{tabular}
\caption{Different encoding strategies for Task 2 (Acc/FNA)}

\end{table}

\vspace{1em}

\begin{table}[H]
\centering
\begin{tabular}{lcccc}
\hline
Model          & Asia & Child & Insurance & Alarm \\
\hline
JSON           & 0.75 & 0.67  & 0.46      & 0.69  \\
GraphML        & 0.71 & 0.67  & 0.50      & 0.72  \\
Textual (ours) & 0.73 & 0.68  & 0.47      & 0.71  \\
\hline
\end{tabular}
\caption{Different encoding strategies for Task 3 (LLM-J)}

\end{table}

~\newpage
\section{Finetuning and Few-shot prompting}
\subsection{Finetuning}
\label{app:ft}
We aim to assess the LLM’s causal reasoning via prompting. The following are the reasons why fine-tuning is not the most practical solution:
\begin{itemize}
    \item Pretrained models come with a wealth of general knowledge, which we aim to leverage. Fine-tuning these models could potentially limit their ability to draw on this broad knowledge base. We aim to understand the utility of pretrained models, as fine-tuning large models like GPT-4 is not always feasible.
    \item The training graph is too small for fine-tuning. Despite considering a large 52-edged graph: Insurance, we would have just 27 datapoints or Alarm with 37 datapoint. Additionally:
\end{itemize}

\begin{enumerate}
    \item Using the same graph as part of train and test would unfortunately lead to training data leakage.
    \item If we consider different graphs for train and test, there would exist a domain shift in the two graphs and the model may be overfitted to the domain of the train graph.
\end{enumerate}

However, to illustrate our hypothesis and alleviate the reviewer's concern, we performed Supervised Fine-Tuning using QLoRA on the Mistral-7b-Instruct model for hypothesizing in the open world task.
The train set here is all of the graphs minus the respective graph it was tested on. We tested on Survey, Insurance and Alzheimers graphs. The model was trained to give one best-fit suggestion for the missing variable.

\begin{table}[!ht]
    \centering
    \begin{tabular}{l|c|c|c}
    \toprule
        ~ & Insurance & Survey & Alzheimers \\ \midrule
        No fine-tuning & 0.42 +- 0.03 & 0.44 +- 0.05 & 0.34 +- 0.04 \\ 
        Fine-tuned & 0.39 +- 0.04 & 0.39 +- 0.03 & 0.36 +- 0.07 \\ \midrule
    \end{tabular}
    \caption{Finetuning results.}
\end{table}
From the above results, it is evident that finetuning does not significantly improve over the prompting results. This is because during training the LLM gets biased towards the domains of training graphs which are contextually distant from the test domain, given the diversity of graphs chosen. One may think that training might help the LLM to understand the task, but from prompt-based model output, it was evident that the LLM can instruction-follow. In summary, we were able to extract the LLM knowledge via prompting and domain-specific fine-tuning could be closely looked at in the future works.

{\subsection{Fewshot prompting}
Similar to fine-tuning, few-shot learning's success depends on balancing domain specificity and generality. To avoid test examples becoming part of the shots, we have to use different domains as examples. Given the complexity of the Alarm graph, we decided to use them as a prior. We performed experiments with 1-shot and 5-shots for the Mixtral 8x7b model. 
\begin{table}[!ht]
    \centering
    \begin{tabular}{l|c|c|c}
    \toprule
    {graph} & {0-shot} & {1-shot} & {5-shot} \\ \midrule
    {Cancer}      & {0.41} & {0.43} & {0.46} \\
    {Survey}      & {0.39} & {0.38} & {0.36} \\
    {Asia}        & {0.66} & {0.70} & {0.72} \\
    {Alzheimer's} & {0.31} & {0.33} & {0.34} \\
    {Child}       & {0.53} & {0.55} & {0.56} \\
    {Insurance}   & {0.46} & {0.42} & {0.45} \\
    \bottomrule
    \end{tabular}
    \caption{Fewshot prompting results.}
    \label{tab:fewshot}
\end{table}
We would like to remind you that Alarm was a medical graph which means that providing more examples in a different domain might hinder the model performance. Drop in performance when changing domain for in-context learning has been discussed in \cite{kwan2024mt} and \cite{gupta2024llm}.
}

\newpage

\section{Confounders}
\label{app:confounders}

\begin{table}[h]
    \centering
    \begin{tabular}{l|c|c|c|c|c|c|c|c|c|c}
    \toprule
        ~ & Sachs & Alarm1 & Alarm2 & Ins1 & Ins2 & Ins3 & Ins4 & Ins5 & Ins6 & Ins7 \\ \midrule
        Zephyr & 0.12 & 0.37 & 0.29 & 0.45 & 0.49 & 0.37 & 0.29 & 0.33 & 0.46 & 0.73 \\ 
        Mixtral & 0.89 & 0.54 & 0.57 & 0.57 & 1.0 & 0.32 & 0.23 & 0.38 & 0.28 & 1.0 \\ 
        Neural & 0.34 & 0.27 & 0.28 & 0.42 & 0.47 & 0.34 & 0.48 & 0.48 & 0.38 & 0.48 \\ 
        LLama & 0.27 & 0.39 & 0.44 & 0.55 & 1.0 & 0.29 & 0.22 & 0.57 & 0.45 & 1.0 \\ 
        Mistral & 0.23 & 0.62 & 0.46 & 0.58 & 1.0 & 0.28 & 0.28 & 0.28 & 0.28 & 1.0 \\ 
        GPT-4 & 0.91 & 0.49 & 0.44 & 0.62 & 0.39 & 0.58 & 0.44 & 0.58 & 0.52 & 1.0 \\ \bottomrule
    \end{tabular}
    \caption{Semantic similarity}
    \label{app:conf1}
\end{table}

\begin{table}[h]
    \centering
    \begin{tabular}{l|c|c|c|c|c|c|c|c|c|c}
    \toprule
        ~ & Sachs & Alarm1 & Alarm2 & Ins1 & Ins2 & Ins3 & Ins4 & Ins5 & Ins6 & Ins7 \\ \midrule
        Zephyr & 0.10 & 0.40 & 0.30 & 0.45 & 0.60 & 0.40 & 0.40 & 0.30 & 0.70 & 0.80 \\ 
        Mixtral & 0.95 & 0.70 & 1.0 & 0.75 & 1.0 & 0.80 & 0.20 & 0.20 & 0.20 & 1.0 \\ 
        Neural & 0.30 & 0.60 & 0.30 & 1.0 & 0.60 & 0.30 & 0.80 & 0.30 & 0.40 & 0.60 \\ 
        LLama & 0.20 & 0.50 & 0.44 & 0.40 & 1.0 & 0.50 & 0.20 & 0.70 & 0.45 & 1.0 \\ 
        Mistral & 0.20 & 0.90 & 0.80 & 0.55 & 1.0 & 0.30 & 0.20 & 0.70 & 0.30 & 1.0 \\ 
        GPT-4 & 0.95 & 0.65 & 0.80 & 0.60 & 0.70 & 0.80 & 0.85 & 0.80 & 0.75 & 1.0 \\ \bottomrule
    \end{tabular}
    \caption{LLM judge}
    \label{app:conf2}
\end{table}

\begin{figure}[h] 
    \centering
    \includegraphics[width=0.5 \textwidth]{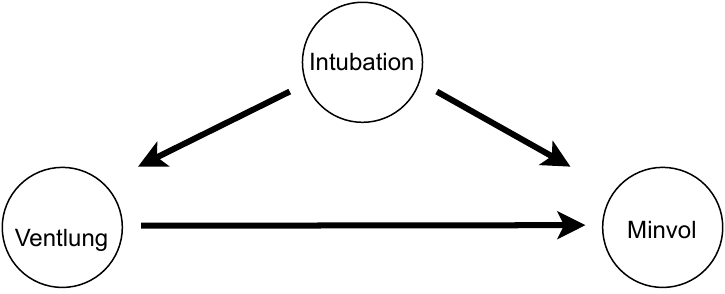}
    \caption{Alarm 1}
    \label{confoundera}
\end{figure}

\begin{figure}[] 
    \centering
    \includegraphics[width=0.5 \textwidth]{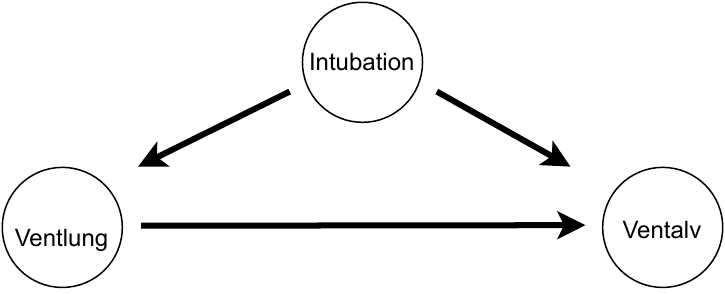}
    \caption{Alarm 2}
    \label{confoundera}
\end{figure}

\begin{figure}[] 
    \centering
    \includegraphics[width=0.5 \textwidth]{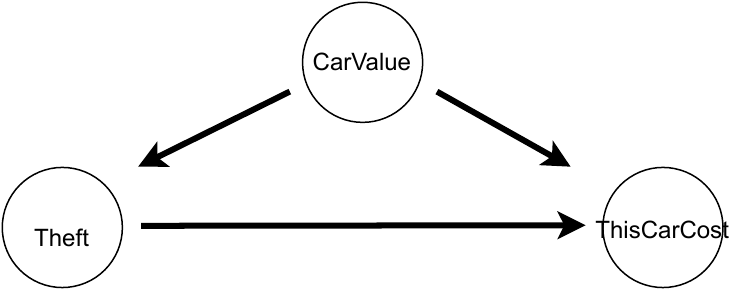}
    \caption{Insurance 1}
    \label{confoundera}
\end{figure}

\begin{figure}[] 
    \centering
    \includegraphics[width=0.5 \textwidth]{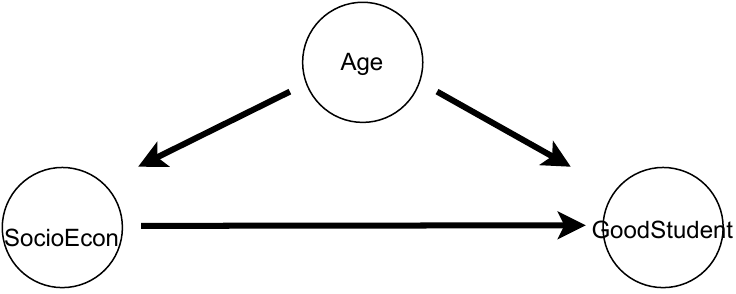}
    \caption{Insurance 2}
    \label{confoundera}
\end{figure}

\begin{figure}[] 
    \centering
    \includegraphics[width=0.5 \textwidth]{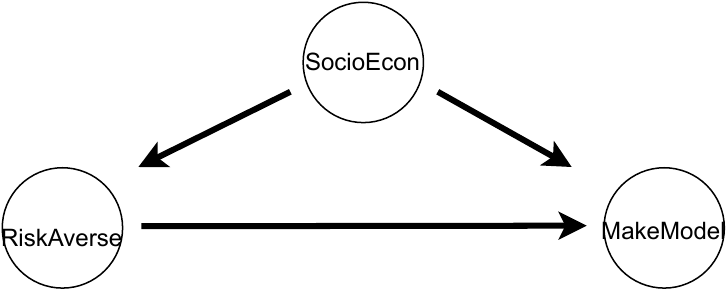}
    \caption{Insurance 3}
    \label{confoundera}
\end{figure}

\begin{figure}[] 
    \centering
    \includegraphics[width=0.5 \textwidth]{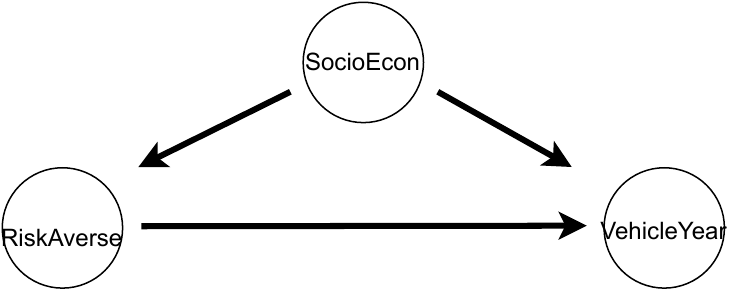}
    \caption{Insurance 4}
    \label{confoundera}
\end{figure}

\begin{figure}[] 
    \centering
    \includegraphics[width=0.5 \textwidth]{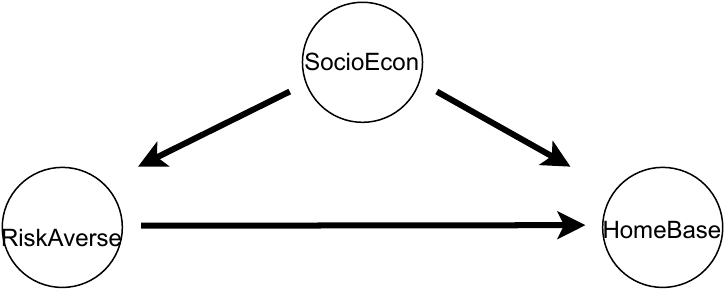}
    \caption{Insurance 5}
    \label{confoundera}
\end{figure}

\begin{figure}[] 
    \centering
    \includegraphics[width=0.5 \textwidth]{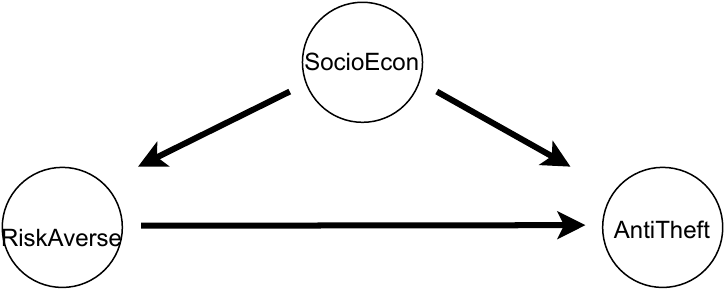}
    \caption{Insurance 6}
    \label{confoundera}
\end{figure}

\begin{figure}[] 
    \centering
    \includegraphics[width=0.5 \textwidth]{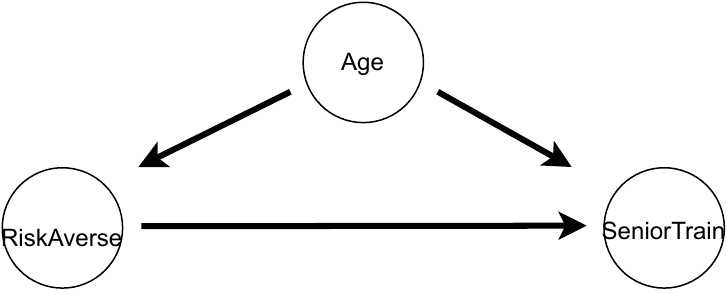}
    \caption{Insurance 7}
    \label{confoundera}
\end{figure}

\section{Causal graphs}

\begin{figure}[H]
\centering
  \begin{tikzpicture}[scale=2]
      \draw
        (0.0:2) node (Cancer){Cancer}
        (72.0:2) node (Dyspnoea){Dyspnoea}
        (144.0:2) node (Pollution){Pollution}
        (216.0:2) node (Smoker){Smoker}
        (288.0:2) node (Xray){Xray};
      \begin{scope}[->]
        \draw (Cancer) to (Xray);
        \draw (Cancer) to (Dyspnoea);
        \draw (Pollution) to (Cancer);
        \draw (Smoker) to (Cancer);
      \end{scope}
    \end{tikzpicture}
    \caption{Cancer DAG}
\end{figure}
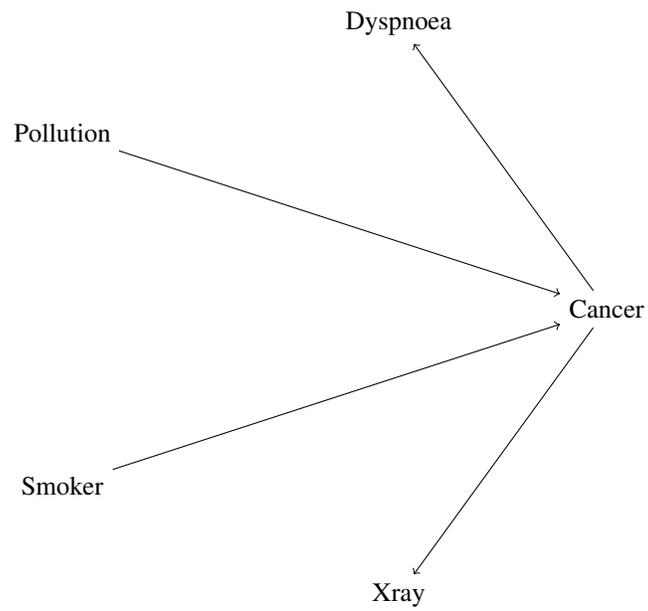

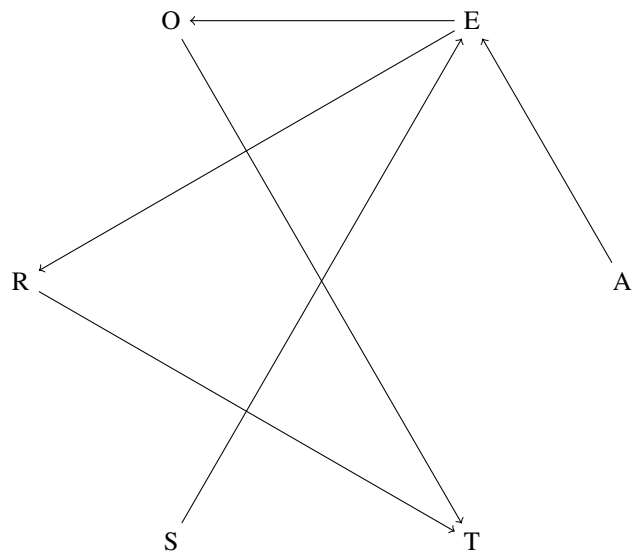
\begin{figure}[H]
\centering
  \begin{tikzpicture}[scale=2]
      \draw
        (0.0:2) node (A){A}
        (60.0:2) node (E){E}
        (120.0:2) node (O){O}
        (180.0:2) node (R){R}
        (240.0:2) node (S){S}
        (300.0:2) node (T){T};
      \begin{scope}[->]
        \draw (A) to (E);
        \draw (E) to (O);
        \draw (E) to (R);
        \draw (O) to (T);
        \draw (R) to (T);
        \draw (S) to (E);
      \end{scope}
    \end{tikzpicture}
    \caption{Survey DAG}
\end{figure}
\newpage

\begin{figure}[H]
\centering
  \begin{tikzpicture}[scale=2]
      \draw
        (0.0:2) node (asia){asia}
        (45.0:2) node (bronc){bronc}
        (90.0:2) node (dysp){dysp}
        (135.0:2) node (either){either}
        (180.0:2) node (lung){lung}
        (225.0:2) node (smoke){smoke}
        (270.0:2) node (tub){tub}
        (315.0:2) node (xray){xray};
      \begin{scope}[->]
        \draw (asia) to (tub);
        \draw (bronc) to (dysp);
        \draw (either) to (xray);
        \draw (either) to (dysp);
        \draw (lung) to (either);
        \draw (smoke) to (lung);
        \draw (smoke) to (bronc);
        \draw (tub) to (either);
      \end{scope}
    \end{tikzpicture}
    \caption{Asia DAG}
\end{figure}
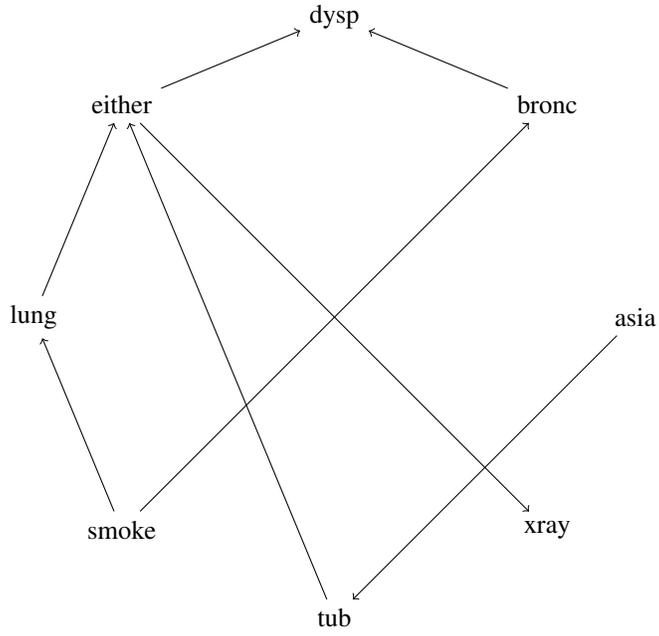

\begin{figure}[H]
\centering
  \begin{tikzpicture}[scale=2]
      \draw
        (0.0:2) node (APOE4){APOE4}
        (40.0:2) node (AV45){AV45}
        (80.0:2) node (Age){Age}
        (120.0:2) node (Brain Volume){Brain Volume}
        (160.0:2) node (Education){Education}
        (200.0:2) node (MOCA Score){MOCA Score}
        (240.0:2) node (Sex){Sex}
        (280.0:2) node (Tau Levels){Tau Levels}
        (320.0:2) node (Ventricular Volume){Ventricular Volume};
      \begin{scope}[->]
        \draw (APOE4) to (AV45);
        \draw (AV45) to (Tau Levels);
        \draw (AV45) to (Brain Volume);
        \draw (Age) to (AV45);
        \draw (Age) to (Tau Levels);
        \draw (Age) to (Brain Volume);
        \draw (Age) to (Ventricular Volume);
        \draw (Age) to (MOCA Score);
        \draw (Brain Volume) to (Ventricular Volume);
        \draw (Brain Volume) to (MOCA Score);
        \draw (Education) to (MOCA Score);
        \draw (Sex) to (Brain Volume);
        \draw (Sex) to (Ventricular Volume);
        \draw (Tau Levels) to (Brain Volume);
        \draw (Tau Levels) to (Ventricular Volume);
        \draw (Tau Levels) to (MOCA Score);
      \end{scope}
    \end{tikzpicture}
    \caption{Alzheimer's DAG}
\end{figure}
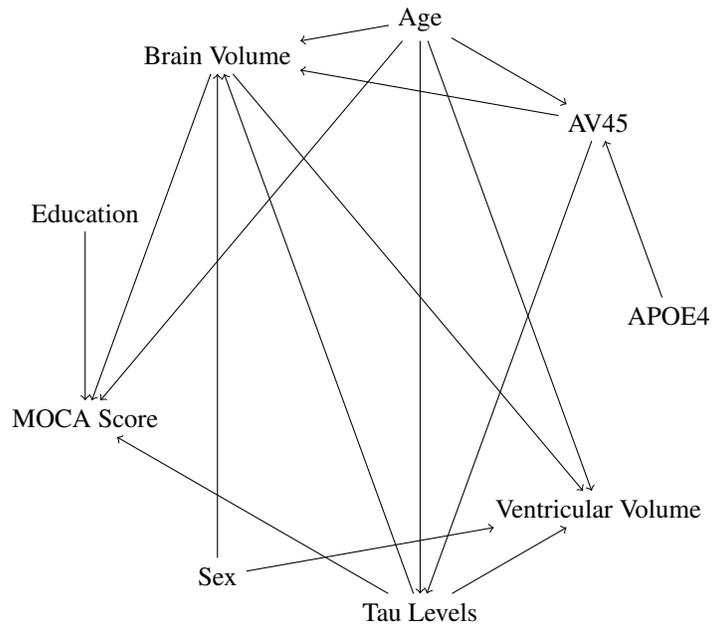

\newpage
\vspace{40mm}
\begin{figure}[H]
\centering
  \begin{tikzpicture}[scale=3.1]
      \draw
        (0.0:2) node (Age){Age}
        (18.0:2) node (BirthAsphyxia){BirthAsphyxia}
        (36.0:2) node (CO2){CO2}
        (54.0:2) node (CO2Report){CO2Report}
        (72.0:2) node (CardiacMixing){CardiacMixing}
        (90.0:2) node (ChestXray){ChestXray}
        (108.0:2) node (Disease){Disease}
        (126.0:2) node (DuctFlow){DuctFlow}
        (144.0:2) node (Grunting){Grunting}
        (162.0:2) node (GruntingReport){GruntingReport}
        (180.0:2) node (HypDistrib){HypDistrib}
        (198.0:2) node (HypoxiaInO2){HypoxiaInO2}
        (216.0:2) node (LVH){LVH}
        (234.0:2) node (LVHreport){LVHreport}
        (252.0:2) node (LowerBodyO2){LowerBodyO2}
        (270.0:2) node (LungFlow){LungFlow}
        (288.0:2) node (LungParench){LungParench}
        (306.0:2) node (RUQO2){RUQO2}
        (324.0:2) node (Sick){Sick}
        (342.0:2) node (XrayReport){XrayReport};
      \begin{scope}[->]
        \draw (BirthAsphyxia) to (Disease);
        \draw (CO2) to (CO2Report);
        \draw (CardiacMixing) to (HypDistrib);
        \draw (CardiacMixing) to (HypoxiaInO2);
        \draw (ChestXray) to (XrayReport);
        \draw (Disease) to (DuctFlow);
        \draw (Disease) to (CardiacMixing);
        \draw (Disease) to (LungParench);
        \draw (Disease) to (LungFlow);
        \draw (Disease) to (Sick);
        \draw (Disease) to (LVH);
        \draw (Disease) to (Age);
        \draw (DuctFlow) to (HypDistrib);
        \draw (Grunting) to (GruntingReport);
        \draw (HypDistrib) to (LowerBodyO2);
        \draw (HypoxiaInO2) to (LowerBodyO2);
        \draw (HypoxiaInO2) to (RUQO2);
        \draw (LVH) to (LVHreport);
        \draw (LungFlow) to (ChestXray);
        \draw (LungParench) to (HypoxiaInO2);
        \draw (LungParench) to (CO2);
        \draw (LungParench) to (ChestXray);
        \draw (LungParench) to (Grunting);
        \draw (Sick) to (Grunting);
        \draw (Sick) to (Age);
      \end{scope}
    \end{tikzpicture}
    \caption{Child DAG}
\end{figure}
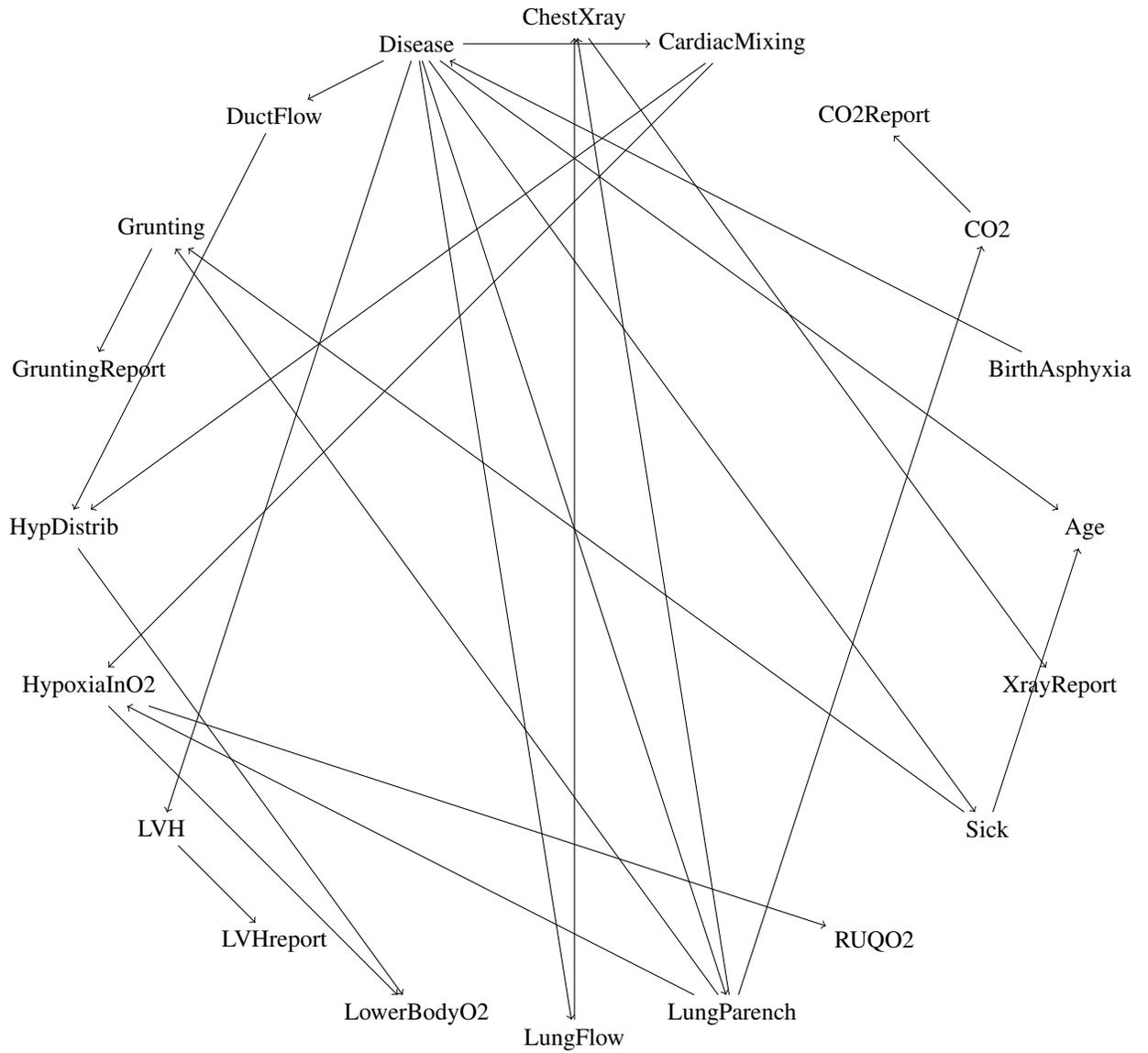

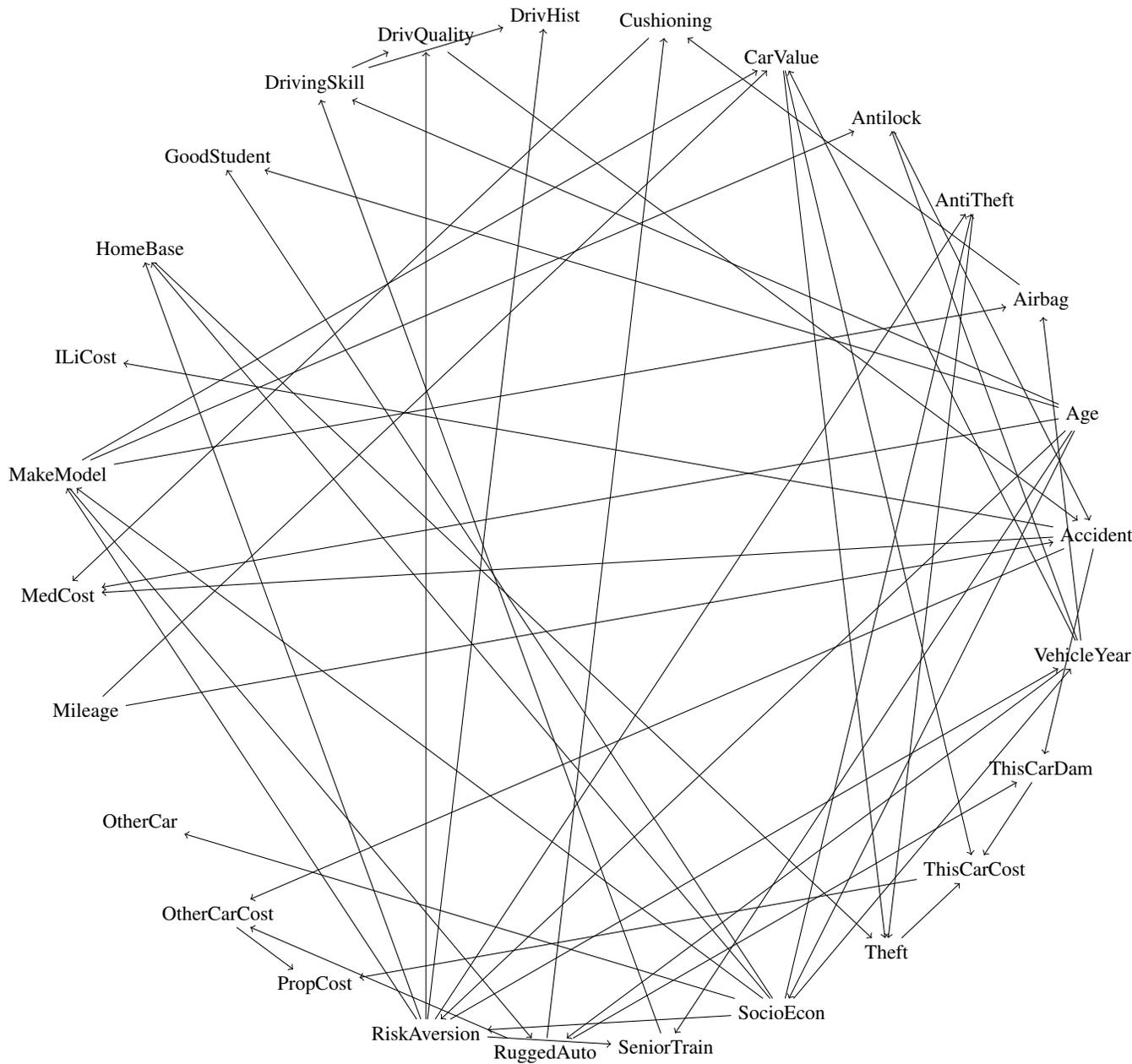
\begin{figure}
\centering
\small
\newpage
\begin{tikzpicture}[scale=3][h]
      \draw
        (0.0:2) node (Accident){Accident}
        (13.333:2) node (Age){Age}
        (26.667:2) node (Airbag){Airbag}
        (40.0:2) node (AntiTheft){AntiTheft}
        (53.333:2) node (Antilock){Antilock}
        (66.667:2) node (CarValue){CarValue}
        (80.0:2) node (Cushioning){Cushioning}
        (93.333:2) node (DrivHist){DrivHist}
        (106.667:2) node (DrivQuality){DrivQuality}
        (120.0:2) node (DrivingSkill){DrivingSkill}
        (133.333:2) node (GoodStudent){GoodStudent}
        (146.667:2) node (HomeBase){HomeBase}
        (160.0:2) node (ILiCost){ILiCost}
        (173.333:2) node (MakeModel){MakeModel}
        (186.667:2) node (MedCost){MedCost}
        (200.0:2) node (Mileage){Mileage}
        (213.333:2) node (OtherCar){OtherCar}
        (226.667:2) node (OtherCarCost){OtherCarCost}
        (240.0:2) node (PropCost){PropCost}
        (253.333:2) node (RiskAversion){RiskAversion}
        (266.667:2) node (RuggedAuto){RuggedAuto}
        (280.0:2) node (SeniorTrain){SeniorTrain}
        (293.333:2) node (SocioEcon){SocioEcon}
        (306.667:2) node (Theft){Theft}
        (320.0:2) node (ThisCarCost){ThisCarCost}
        (333.333:2) node (ThisCarDam){ThisCarDam}
        (346.667:2) node (VehicleYear){VehicleYear};
      \begin{scope}[->]
        \draw (Accident) to (ThisCarDam);
        \draw (Accident) to (OtherCarCost);
        \draw (Accident) to (MedCost);
        \draw (Accident) to (ILiCost);
        \draw (Age) to (SocioEcon);
        \draw (Age) to (GoodStudent);
        \draw (Age) to (RiskAversion);
        \draw (Age) to (DrivingSkill);
        \draw (Age) to (SeniorTrain);
        \draw (Age) to (MedCost);
        \draw (Airbag) to (Cushioning);
        \draw (AntiTheft) to (Theft);
        \draw (Antilock) to (Accident);
        \draw (CarValue) to (ThisCarCost);
        \draw (CarValue) to (Theft);
        \draw (Cushioning) to (MedCost);
        \draw (DrivQuality) to (Accident);
        \draw (DrivingSkill) to (DrivQuality);
        \draw (DrivingSkill) to (DrivHist);
        \draw (HomeBase) to (Theft);
        \draw (MakeModel) to (RuggedAuto);
        \draw (MakeModel) to (Antilock);
        \draw (MakeModel) to (CarValue);
        \draw (MakeModel) to (Airbag);
        \draw (Mileage) to (Accident);
        \draw (Mileage) to (CarValue);
        \draw (OtherCarCost) to (PropCost);
        \draw (RiskAversion) to (VehicleYear);
        \draw (RiskAversion) to (MakeModel);
        \draw (RiskAversion) to (DrivQuality);
        \draw (RiskAversion) to (SeniorTrain);
        \draw (RiskAversion) to (AntiTheft);
        \draw (RiskAversion) to (HomeBase);
        \draw (RiskAversion) to (DrivHist);
        \draw (RuggedAuto) to (ThisCarDam);
        \draw (RuggedAuto) to (OtherCarCost);
        \draw (RuggedAuto) to (Cushioning);
        \draw (SeniorTrain) to (DrivingSkill);
        \draw (SocioEcon) to (GoodStudent);
        \draw (SocioEcon) to (RiskAversion);
        \draw (SocioEcon) to (VehicleYear);
        \draw (SocioEcon) to (MakeModel);
        \draw (SocioEcon) to (AntiTheft);
        \draw (SocioEcon) to (HomeBase);
        \draw (SocioEcon) to (OtherCar);
        \draw (Theft) to (ThisCarCost);
        \draw (ThisCarCost) to (PropCost);
        \draw (ThisCarDam) to (ThisCarCost);
        \draw (VehicleYear) to (RuggedAuto);
        \draw (VehicleYear) to (Antilock);
        \draw (VehicleYear) to (CarValue);
        \draw (VehicleYear) to (Airbag);
      \end{scope}
    \end{tikzpicture}
    \caption{Insurance DAG}
\end{figure}

\begin{figure}[H]
\centering
  \begin{tikzpicture}[scale=3.0]
  \small
      \draw
        (0.0:2) node (ANAPHYLAXIS){ANAPHYLAXIS}
        (9.73:2) node (ARTCO2){ARTCO2}
        (19.459:2) node (BP){BP}
        (29.189:2) node (CATECHOL){CATECHOL}
        (38.919:2) node (CO){CO}
        (48.649:2) node (CVP){CVP}
        (58.378:2) node (DISCONNECT){DISCONNECT}
        (68.108:2) node (ERRCAUTER){ERRCAUTER}
        (77.838:2) node (ERRLOWOUTPUT){ERRLOWOUTPUT}
        (87.568:2) node (EXPCO2){EXPCO2}
        (97.297:2) node (FIO2){FIO2}
        (107.027:2) node (HISTORY){HISTORY}
        (116.757:2) node (HR){HR}
        (126.486:2) node (HRBP){HRBP}
        (136.216:2) node (HREKG){HREKG}
        (145.946:2) node (HRSAT){HRSAT}
        (155.676:2) node (HYPOVOLEMIA){HYPOVOLEMIA}
        (165.405:2) node (INSUFFANESTH){INSUFFANESTH}
        (175.135:2) node (INTUBATION){INTUBATION}
        (184.865:2) node (KINKEDTUBE){KINKEDTUBE}
        (194.595:2) node (LVEDVOLUME){LVEDVOLUME}
        (204.324:2) node (LVFAILURE){LVFAILURE}
        (214.054:2) node (MINVOL){MINVOL}
        (223.784:2) node (MINVOLSET){MINVOLSET}
        (233.514:2) node (PAP){PAP}
        (243.243:2) node (PCWP){PCWP}
        (252.973:2) node (PRESS){PRESS}
        (262.703:2) node (PULMEMBOLUS){PULMEMBOLUS}
        (272.432:2) node (PVSAT){PVSAT}
        (282.162:2) node (SAO2){SAO2}
        (291.892:2) node (SHUNT){SHUNT}
        (301.622:2) node (STROKEVOLUME){STROKEVOLUME}
        (311.351:2) node (TPR){TPR}
        (321.081:2) node (VENTALV){VENTALV}
        (330.811:2) node (VENTLUNG){VENTLUNG}
        (340.541:2) node (VENTMACH){VENTMACH}
        (350.27:2) node (VENTTUBE){VENTTUBE};
      \begin{scope}[->]
        \draw (ANAPHYLAXIS) to (TPR);
        \draw (ARTCO2) to (EXPCO2);
        \draw (ARTCO2) to (CATECHOL);
        \draw (CATECHOL) to (HR);
        \draw (CO) to (BP);
        \draw (DISCONNECT) to (VENTTUBE);
        \draw (ERRCAUTER) to (HREKG);
        \draw (ERRCAUTER) to (HRSAT);
        \draw (ERRLOWOUTPUT) to (HRBP);
        \draw (FIO2) to (PVSAT);
        \draw (HR) to (HRBP);
        \draw (HR) to (HREKG);
        \draw (HR) to (HRSAT);
        \draw (HR) to (CO);
        \draw (HYPOVOLEMIA) to (LVEDVOLUME);
        \draw (HYPOVOLEMIA) to (STROKEVOLUME);
        \draw (INSUFFANESTH) to (CATECHOL);
        \draw (INTUBATION) to (VENTLUNG);
        \draw (INTUBATION) to (MINVOL);
        \draw (INTUBATION) to (VENTALV);
        \draw (INTUBATION) to (SHUNT);
        \draw (INTUBATION) to (PRESS);
        \draw (KINKEDTUBE) to (VENTLUNG);
        \draw (KINKEDTUBE) to (PRESS);
        \draw (LVEDVOLUME) to (CVP);
        \draw (LVEDVOLUME) to (PCWP);
        \draw (LVFAILURE) to (HISTORY);
        \draw (LVFAILURE) to (LVEDVOLUME);
        \draw (LVFAILURE) to (STROKEVOLUME);
        \draw (MINVOLSET) to (VENTMACH);
        \draw (PULMEMBOLUS) to (SHUNT);
        \draw (PULMEMBOLUS) to (PAP);
        \draw (PVSAT) to (SAO2);
        \draw (SAO2) to (CATECHOL);
        \draw (SHUNT) to (SAO2);
        \draw (STROKEVOLUME) to (CO);
        \draw (TPR) to (CATECHOL);
        \draw (TPR) to (BP);
        \draw (VENTALV) to (ARTCO2);
        \draw (VENTALV) to (PVSAT);
        \draw (VENTLUNG) to (EXPCO2);
        \draw (VENTLUNG) to (MINVOL);
        \draw (VENTLUNG) to (VENTALV);
        \draw (VENTMACH) to (VENTTUBE);
        \draw (VENTTUBE) to (VENTLUNG);
        \draw (VENTTUBE) to (PRESS);
      \end{scope}
    \end{tikzpicture}
    \caption{Alarm DAG}
\end{figure}
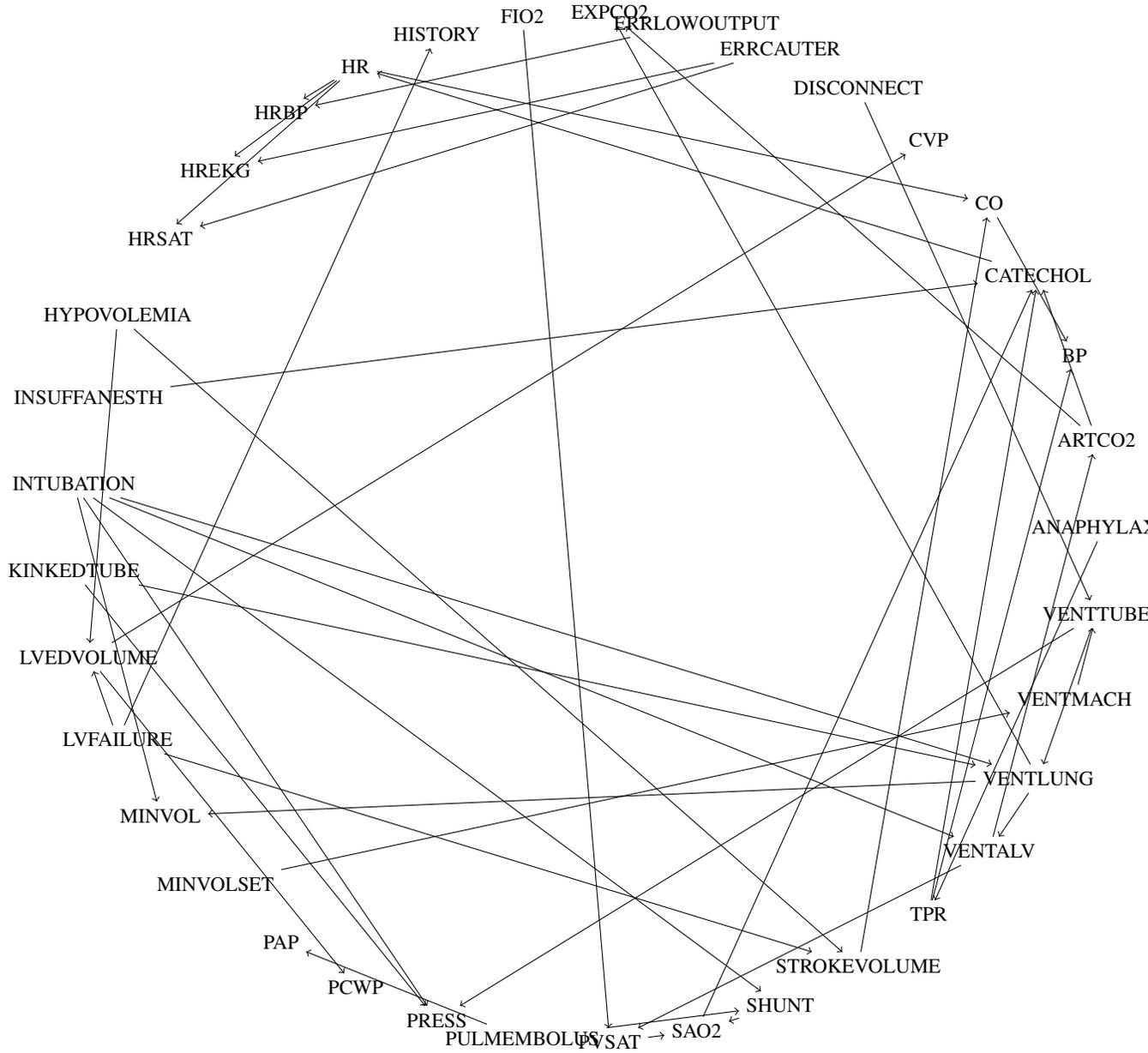

\newpage

\section{Prompt template}
\label{app:prompt}
\begin{PROMPT}
    \centering
    \begin{minipage}{0.9\linewidth}  

    \fbox{
    \parbox{\columnwidth}{
    Hello. You will be given a causal graph. The context of the graph [CONTEXT]. Please understand the causal relationships between the variables - [VERBALISED DAG]. 
    }}
    \caption{Base prompt to describe the causal graph}
    \label{prompt:}
    \end{minipage}
\end{PROMPT}

\begin{PROMPT} 
    \centering
        \begin{minipage}{0.9\linewidth}  

    \fbox{
    \parbox{\columnwidth}{
     Hello. You will be given a causal graph. The context of the graph is hypothetical patient monitoring system in an intensive care unit (ICU). Please understand the causal relationships between the variables - $<$ anaphylaxis $>$ causes $<$ total peripheral resistance $>$. $<$ arterial co2 $>$ causes $<$ expelled co2 $>$. $<$ arterial co2 $>$ causes $<$ catecholamine $>$. $<$ catecholamine $>$ causes $<$ heart rate $>$. $<$ cardiac output $>$ causes $<$ blood pressure $>$. $<$ disconnection $>$ causes $<$ breathing tube $>$. $<$ error cauter $>$ causes $<$ heart rate displayed on ekg monitor $>$. $<$ error cauter $>$ causes $<$ oxygen saturation $>$. $<$ error low output $>$ causes $<$ heart rate blood pressure $>$. $<$ high concentration of oxygen in the gas mixture  $>$ causes $<$ pulmonary artery oxygen saturation $>$. $<$ heart rate $>$ causes $<$ heart rate blood pressure $>$. $<$ heart rate $>$ causes $<$ heart rate displayed on ekg monitor $>$. $<$ heart rate $>$ causes $<$ oxygen saturation $>$. $<$ heart rate $>$ causes $<$ cardiac output $>$. $<$ hypovolemia $>$ causes $<$ left ventricular end-diastolic volume $>$. $<$ hypovolemia $>$ causes $<$ stroke volume  $>$. $<$ insufficient anesthesia  $>$ causes $<$ catecholamine $>$. $<$ intubation $>$ causes $<$ lung ventilation $>$. $<$ intubation $>$ causes $<$ minute volume  $>$. $<$ intubation $>$ causes $<$ alveolar ventilation $>$. $<$ intubation $>$ causes $<$ shunt - normal and high $>$. $<$ intubation $>$ causes $<$ breathing pressure $>$. $<$ kinked chest tube $>$ causes $<$ lung ventilation $>$. $<$ kinked chest tube $>$ causes $<$ breathing pressure $>$. $<$ left ventricular end-diastolic volume $>$ causes $<$ central venous pressure $>$. $<$ left ventricular end-diastolic volume $>$ causes $<$ pulmonary capillary wedge pressure $>$. $<$ left ventricular failure $>$ causes $<$ previous medical history $>$. $<$ left ventricular failure $>$ causes $<$ left ventricular end-diastolic volume $>$. $<$ left ventricular failure $>$ causes $<$ stroke volume  $>$. $<$ the amount of time using a breathing machine $>$ causes $<$ the intensity level of a breathing machine $>$. $<$ sudden blockage in the pulmonary arteries $>$ causes $<$ shunt - normal and high $>$. $<$ sudden blockage in the pulmonary arteries $>$ causes $<$ pulmonary artery pressure  $>$. $<$ pulmonary artery oxygen saturation $>$ causes $<$ oxygen saturation $>$. $<$ oxygen saturation $>$ causes $<$ catecholamine $>$. $<$ shunt - normal and high $>$ causes $<$ oxygen saturation $>$. $<$ stroke volume  $>$ causes $<$ cardiac output $>$. $<$ total peripheral resistance $>$ causes $<$ catecholamine $>$. $<$ total peripheral resistance $>$ causes $<$ blood pressure $>$. $<$ alveolar ventilation $>$ causes $<$ arterial co2 $>$. $<$ alveolar ventilation $>$ causes $<$ pulmonary artery oxygen saturation $>$. $<$ lung ventilation $>$ causes $<$ expelled co2 $>$. $<$ lung ventilation $>$ causes $<$ minute volume  $>$. $<$ lung ventilation $>$ causes $<$ alveolar ventilation $>$. $<$ the intensity level of a breathing machine $>$ causes $<$ breathing tube $>$. $<$ breathing tube $>$ causes $<$ lung ventilation $>$. $<$ breathing tube $>$ causes $<$ breathing pressure $>$. 
    }}
    
\caption{An example of the base prompt for Alarm graph. Each relationship is enclosed in pointed brackets,$<>$ followed by a full stop.} 
\label{prompt:}
\end{minipage}
\end{PROMPT}

\clearpage

\begin{PROMPT} 
    \centering
        \begin{minipage}{0.9\linewidth}  

    \fbox{
    \parbox{\columnwidth}{
    \textbf{Prompt:} Hello. You will be given a causal graph. The context of the graph is different aspects of a patient's health related to lung diseases. Please understand the causal relationships between the variables - $<$ x $>$ causes $<$ tuberculosis $>$. $<$ bronchitis $>$ causes $<$ dyspnoea-laboured breathing $>$. $<$ individual has either tuberculosis or lung cancer $>$ causes $<$ positive xray $>$. $<$ individual has either tuberculosis or lung cancer $>$ causes $<$ dyspnoea-laboured breathing $>$. $<$ lung cancer $>$ causes $<$ individual has either tuberculosis or lung cancer $>$. $<$ smoking cigarettes $>$ causes $<$ lung cancer $>$. $<$ smoking cigarettes $>$ causes $<$ bronchitis $>$. $<$ tuberculosis $>$ causes $<$ individual has either tuberculosis or lung cancer $>$.Based on your understanding and analytical skills, please identify what the variable X could most likely be from the given multiple choices. It's important to note that you MUST select an option from the provided choices, as one of them is the correct answer. The choices are: [ visited Asia; weather; book sales; movie rating ]. Please think step by step. Your answer should be formatted as follows - Answer: X = choice. The choice must be one of the provided options, and it should be formatted exactly as it appears in the choices. Generating your own options is not allowed in this task.

\vspace{5mm}

    \textbf{Answer}: X = \textcolor{green}{visited Asia}.My thought process is as follows:The causal graph suggests that X causes tuberculosis, and tuberculosis is related to lung diseases. Visiting Asia is a possible explanation for why an individual may have been exposed to tuberculosis, as the disease is more prevalent in some Asian countries. Therefore, visited Asia is the most likely option for X.
}}
\caption{Out-of-context controlled variable identification, Ground truth variable: visited Asia} 
\label{prompt:}

\end{minipage}

\end{PROMPT}

\vspace{50mm}

\begin{PROMPT} 
    \centering
        \begin{minipage}{0.9\linewidth}  

    \fbox{
    \parbox{\columnwidth}{
    \textbf{Prompt:} Hello. You will be given a causal graph. The context of the graph is different aspects of a patient's health related to lung diseases Please understand the causal relationships between the variables - $<$ x $>$ causes $<$ tuberculosis $>$. $<$ bronchitis $>$ causes $<$ dyspnoea-laboured breathing $>$. $<$ individual has either tuberculosis or lung cancer $>$ causes $<$ y $>$. $<$ individual has either tuberculosis or lung cancer $>$ causes $<$ dyspnoea-laboured breathing $>$. $<$ lung cancer $>$ causes $<$ individual has either tuberculosis or lung cancer $>$. $<$ smoking cigarettes $>$ causes $<$ lung cancer $>$. $<$ smoking cigarettes $>$ causes $<$ bronchitis $>$. $<$ tuberculosis $>$ causes $<$ individual has either tuberculosis or lung cancer $>$. Based on your understanding and analytical skills, please identify what the variable X could most likely be from the given multiple choices. It's important to note that you MUST select an option from the provided choices, as one of them is the correct answer. The choices are: [ weather; book sales; movie rating; positive xray; visited Asia ]. Please think step by step. Your answer should be formatted as follows - Answer: X = choice. The choice must be one of the provided options, and it should be formatted exactly as it appears in the choices. Generating your own options is not allowed in this task.

\vspace{5mm}

    \textbf{Answer}: Answer: X = \textcolor{green}{visited Asia}. My thought process is as follows:1. The graph shows that X causes tuberculosis, which suggests that X is a variable related to healthcare or medicine. 2. The graph also shows that bronchitis causes dyspnoea-laboured breathing, which further supports the idea that X is a variable related to healthcare or medicine. 3. The graph shows that individual has either tuberculosis or lung cancer causes y, which suggests that X could be a variable related to diagnosing or treating lung diseases. 4. The graph shows that smoking cigarettes causes lung cancer and bronchitis, which suggests that X could be a variable related to smoking or lung health. 5. Based on the above observations, it seems unlikely that X would be weather, book sales, or movie rating.6. Visited Asia is a variable that could potentially relate to healthcare or medicine, particularly if the patient has traveled to a region with a high prevalence of tuberculosis or other lung diseases.7. Therefore, based on the information provided in the graph, it seems most likely that X = \textcolor{green}{visited Asia}.
}}
\caption{In-context controlled variable identification, Ground truth variable: visited asia} 
\label{prompt:}
\end{minipage}
\end{PROMPT}

\clearpage

\begin{PROMPT} 
    \centering
        \begin{minipage}{0.9\linewidth}  

    \fbox{
    \parbox{\columnwidth}{
    \textbf{Prompt:} Hello. You will be given a causal graph. The context of the graph is different aspects of a patient's health related to lung diseases. Please understand the causal relationships between the variables - $<$ x $>$ causes $<$ tuberculosis $>$. $<$ bronchitis $>$ causes $<$ dyspnoea-laboured breathing $>$. $<$ individual has either tuberculosis or lung cancer $>$ causes $<$ positive xray $>$. $<$ individual has either tuberculosis or lung cancer $>$ causes $<$ dyspnoea-laboured breathing $>$. $<$ lung cancer $>$ causes $<$ individual has either tuberculosis or lung cancer $>$. $<$ smoking cigarettes $>$ causes $<$ lung cancer $>$. $<$ smoking cigarettes $>$ causes $<$ bronchitis $>$. $<$ tuberculosis $>$ causes $<$ individual has either tuberculosis or lung cancer $>$. Using your causal knowledge, can you suggest what variable X most likely? Give 5 suggestions for what variable X could be. Give reason for your suggestions. Finally, return your answer (without reasoning) in the following format: $<$Answer$>$ [first suggestion, second suggestion, third suggestion .. and so on] $<$/Answer$>$ No explanations, no reasoning, just answers as a list as described earlier. For example, if the suggestions are A,B,C,D,E then - $<$Answer$>$ [A, B, C, D, E] $<$/Answer$>$

\vspace{5mm}

    \textbf{Answer}: $<$Answer$>$ [HIV/AIDS, Chronic respiratory conditions, \textbf{\textcolor{green}{Pneumonia}}, Exposure to asbestos, Silicosis] $<$/Answer$>$ 
    
    Reasons: 
    
    1. HIV/AIDS: It is known to weaken the immune system, making the individual more susceptible to infections like tuberculosis.
    
    2. Chronic respiratory conditions: These can make an individual more susceptible to lung diseases like tuberculosis.
    
    3. \textbf{\textcolor{green}{Pneumonia}}: It is a lung infection that can lead to serious conditions like tuberculosis if not treated properly.
    
    4. Exposure to asbestos: Long-term exposure to asbestos can lead to lung diseases including lung cancer and tuberculosis.
    
    5. Silicosis: It is a lung disease caused by inhalation of silica dust, leading to inflammation and scarring of the lung tissue, making the individual more susceptible to tuberculosis.
}}
\caption{Hypothesizing missing variable in open world, Ground truth variable: Visited Asia} 
\label{prompt:}
\end{minipage}
\end{PROMPT}

\begin{PROMPT} 
    \centering
        \begin{minipage}{0.9\linewidth}  

    \fbox{
    \parbox{\columnwidth}{
    \textbf{Prompt:} Hello. You will be given a causal graph. The context of the graph is different aspects of a patient's health related to lung diseases. Please understand the causal relationships between the variables - $<$ visited asia $>$ causes $<$ tuberculosis $>$. $<$ bronchitis $>$ causes $<$ dyspnoea-laboured breathing $>$. $<$ individual has either tuberculosis or lung cancer $>$ causes $<$ positive xray $>$. $<$ individual has either tuberculosis or lung cancer $>$ causes $<$ dyspnoea-laboured breathing $>$. $<$ x $>$ causes $<$ individual has either tuberculosis or lung cancer $>$. $<$ smoking cigarettes $>$ causes $<$ x $>$. $<$ smoking cigarettes $>$ causes $<$ bronchitis $>$. $<$ tuberculosis $>$ causes $<$ individual has either tuberculosis or lung cancer $>$. Using your causal knowledge, can you suggest what variable X most likely? Give 5 suggestions for what variable X could be. Give reason for your suggestions. Finally, return your answer (without reasoning) in the following format: $<$Answer$>$ [first suggestion, second suggestion, third suggestion .. and so on] $<$/Answer$>$ No explanations, no reasoning, just answers as a list as described earlier. For example, if the suggestions are A,B,C,D,E then - $<$Answer$>$ [A, B, C, D, E] $<$/Answer$>$

\vspace{5mm}

    \textbf{Answer}: $<$Answer$>$ [\textbf{\textcolor{green}{Lung Cancer}}, Chronic Obstructive Pulmonary Disease, Emphysema, Pneumonia, Asthma] $<$/Answer$>$ 
    
    Reasons: 
    
    1. \textbf{\textcolor{green}{Lung Cancer}}: Smoking is a major cause of lung cancer and lung cancer can lead to tuberculosis or other lung diseases.
    
    2. Chronic Obstructive Pulmonary Disease: Smoking is a leading cause of COPD and it can lead to other lung diseases including tuberculosis and lung cancer.
    
    3. Emphysema: It is a type of COPD often caused by smoking and can lead to other lung diseases.
    
    4. Pneumonia: Smoking increases the risk of pneumonia and pneumonia can lead to other lung diseases.
    
    5. Asthma: Smoking can trigger asthma attacks and severe asthma can lead to other lung diseases.
}}
\caption{Hypothesizing missing variable in open world, Ground truth variable: Lung cancer} 
\label{prompt:}
\end{minipage}
\end{PROMPT}

\clearpage

\newpage
\section{Assumptions}
The \textbf{causal sufficiency} of $\mathcal{G}$, by definition, implies that for every pair of variables within $\mathbf{V}$, all common causes are also included within $\mathbf{V}$. Extending this assumption to $\mathcal{G}^*$, we assume that the partial graph inherits causal sufficiency for its given that all edges among these variables are preserved as in $\mathcal{G}$. This preservation ensures that the observed relationships within $V^*$ are not confounded by omitted common causes. Since the faithfulness of $\mathcal{G}$ ensures that the observed conditional independencies among variables in $\mathbf{V}$ are accurately reflected by the causal structure represented by $\mathbf{E}$. By maintaining the same set of edges $\mathbf{E}$ in $\mathcal{G}^*$ for the subset $V^*$, we uphold the faithfulness assumption within the partial graph.

\end{document}